\documentclass[oneside, a4paper, 12pt, bibliography=totoc]{scrartcl}
\usepackage[a4paper]{anysize}
\usepackage{marginnote}
\usepackage{times}
\usepackage[hyphens]{url}
\usepackage{hyperref}
\usepackage[round,colon]{natbib}
\usepackage[utf8]{inputenc}
\usepackage[fleqn]{amsmath}
\usepackage{amsfonts}
\usepackage{amssymb}
\usepackage{mleftright}
\usepackage{amsthm}
\usepackage{thmtools}
\usepackage{mathtools}
\usepackage{graphicx}
\usepackage{calc}

\begin{document}
\setlength{\mathindent}{0mm}
\newlength{\symaeqmargin}
\setlength{\symaeqmargin}{2cm}
\newlength{\symaeqwidth}
\setlength{\symaeqwidth}{\textwidth - \symaeqmargin}
\allowdisplaybreaks

\newcommand{\mb}[1]{\mathbf{#1}}
\newcommand{\bs}[1]{\boldsymbol{#1}}
\newcommand{\opnl}[1]{\operatorname{#1}\nolimits}
\setlength\nulldelimiterspace{0pt}
%
\RedeclareSectionCommand[tocnumwidth=35pt]{subsection}
\title{A Lagrange-Newton Approach\\to Smoothing-and-Mapping}

\author{Ralf M\"oller\\Computer Engineering, Faculty of
Technology\\Bielefeld University\\\url{www.ti.uni-bielefeld.de}}

\date{\hspace*{1cm}}

\maketitle

\begin{abstract}

\noindent In this report we explore the application of the
Lagrange-Newton method to the SAM (smoothing-and-mapping) problem in
mobile robotics. In Lagrange-Newton SAM, the angular component of each
pose vector is expressed by orientation vectors and treated through
Lagrange constraints. This is different from the typical Gauss-Newton
approach where variations need to be mapped back and forth between
Euclidean space and a manifold suitable for rotational components. We
derive equations for five different types of measurements between robot
poses: translation, distance, and rotation from odometry in the plane,
as well as home-vector angle and compass angle from visual homing. We
demonstrate the feasibility of the Lagrange-Newton approach for a simple
example related to a cleaning robot scenario.

\end{abstract}

\tableofcontents

\section{Introduction}\label{sec_intro}

%
Smoothing-and-mapping (SAM) is a fairly modern approach to building
graph-based maps of the environment; a tutorial is given by
\citet[]{nav_Grisetti10} (see also the tutorial on Newton-type methods
with applications in graph-based SLAM by \citet[]{nn_Toussaint17}).
Following the tutorial by \citet[]{nav_Grisetti10}, the SAM problem (1)
is solved in a Gauss-Newton framework and (2) involves transformations
between manifolds for the rotational components. In the Gauss-Newton
framework, error vectors (between true and expected measurements) are
approximated to first order by a Taylor expansion, thus the resulting
scalar error terms are second-order equations. A Newton descent is then
performed based on the approximated Hessian (computed from the Jacobian)
in these equations.

However, while the Gauss-Newton framework allows for straightforward
derivations, it suffers from a drawback with respect to rotational cost
functions: In the quadratic approximation of the scalar error functions,
the cyclic character of angular terms is lost. Therefore, to handle
rotational components (e.g. pose angles), variations need to be
transformed between the Euclidean space (assumed by the Gauss-Newton
approach) and a manifold suitable for the rotational components
\citep[][]{nav_Grisetti10}. Deriving and understanding these
transformations is demanding.

Here we explore a different approach to the SAM problem where we (1)
perform a full Newton descent and (2) treat rotational components by
constraints in a Lagrange-Newton framework. For the Newton descent, we
compute the exact Hessian, not an approximation. Rotational components
are described by orientation vectors with the unit-length constraint
expressed by Lagrange constraints. Lagrange multipliers become part of
the state vector. Note that a Newton descent is not only faster than a
gradient descent, but indispensable if the Lagrange multipliers are
included since in this case the extreme point is always a saddle;
multiplication of the gradient by the inverse Hessian turns the saddle
into an attractor.

The core idea explored here is to express rotational components through
orientation vectors (with unit length) instead of angles. In this way,
the problem space remains Euclidean. Equality constraints expressing the
unit-length property are handled by the Lagrange-multiplier method (2).
The fact that we apply Newton's method with an exact computation of the
Hessian (1) is currently a necessity since we didn't see whether and how
the Gauss-Newton framework would be compatible with this idea. It is
also not clear whether computing the exact Hessian would improve the
performance of the method compared to the approximated Hessian. An
obvious disadvantage is the preparatory effort required to symbolically
compute the exact Hessian. Moreover, the algorithmic solution of the
Lagrange-Newton iteration is complicated by the saddle-point property
mentioned above (e.g. for choosing a criterion for the line-search).

We apply our Lagrange-Newton approach to the problem of navigation in
the plane. We define five cost functions: translation, distance, and
rotation error derived from an odometry model, and home-vector and
compass error related to a visual homing method. We also explore
different variants of the three rotational cost functions (rotation,
home-vector, and compass error). The visual homing method determines two
estimates from a pair of panoramic views captured at two different
locations: the direction (home vector) from one capture point to the
other, and the azimuthal rotation between the views (compass). This is
the approach used in our previous work on cleaning robot navigation
\citep[][]{own_Moeller13,own_Hillen13}. Also our exploratory simulation
is related to the cleaning-robot scenario.

After introducing the notation in section \ref{sec_notation}, we define
planar orientation vectors which express angular terms and explore their
properties in section \ref{sec_orvec}. The five cost functions are
defined in section \ref{sec_cost}. The first- and second-order
derivatives required for the Newton method are determined in section
\ref{sec_deriv}. All derivatives are combined in the terms required for
the Lagrange-Newton descent in section \ref{sec_lagrange_newton}.
Section \ref{sec_exp} describes the preliminary experiments and their
results. Conclusions are presented in section \ref{sec_conclusions}.

\section{Notation}\label{sec_notation}

%
The two poses paired in a measurement are called ``first point/pose''
and ``second point/pose'' here.
\textbf{Dimensions}
\begin{description}
\setlength{\itemsep}{0pt}
\item[] $N$: number of poses, $2 \leq N$
\item[] $M_o$: number of odometry measurements (translation, distance, rotation), $1 \leq M_o$
\item[] $M_h$: number of homing measurements (home vector, compass), $1 \leq M_h$
\end{description}
\textbf{Odometry Measurement}
\begin{description}
\setlength{\itemsep}{0pt}
\item[] $\mb{C}$: odometry covariance (in robot coordinates of first point), symmetric, $3 \times 3$
\item[] $\mb{T}$: translatory covariance (in robot coordinates of first point), symmetric, $2 \times 2$
\item[] $\sigma$: angular standard deviation of odometry, scalar, $1 \times 1$
\item[] $\sigma_e$: standard deviation of odometry distance, scalar, $1 \times 1$
\item[] $\mb{b}$: cross-covariance terms in odometry covariance matrix, arbitrary, $2 \times 1$
\item[] $\mb{r}$: odometry translation vector (in robot coordinates of first point), arbitrary, $2 \times 1$
\item[] $\varrho$: odometry distance (length of translation vector), scalar, $1 \times 1$
\item[] $\mb{q}$: odometry orientation change vector, arbitrary, $2 \times 1$
\item[] $\mb{Q}$: odometry orientation change matrix, orthogonal, $2 \times 2$
\end{description}
\textbf{Homing Measurement}
\begin{description}
\setlength{\itemsep}{0pt}
\item[] $\bs{\alpha}$: home orientation vector (in robot coordinates), arbitrary, $2 \times 1$
\item[] $\bs{A}$: home orientation matrix (in robot coordinates), orthogonal, $2 \times 2$
\item[] $\sigma_h$: standard deviation of home-vector measurement, scalar, $1 \times 1$
\item[] $\bs{\psi}$: compass orientation vector (in robot coordinates), arbitrary, $2 \times 1$
\item[] $\bs{\Psi}$: compass orientation matrix (in robot coordinates), orthogonal, $2 \times 2$
\item[] $\sigma_c$: standard deviation of compass measurement, scalar, $1 \times 1$
\end{description}
\textbf{Pose Estimates}
\begin{description}
\setlength{\itemsep}{0pt}
\item[] $\bs{\delta}$: position change (estimate, in world coordinates), arbitrary, $2 \times 1$
\item[] $\bs{\delta}_0$: unit vector of $\bs{\delta}$, arbitrary, $2 \times 1$
\item[] $\mb{d}$: position change (estimate, in robot coordinates of first point), arbitrary, $2 \times 1$
\item[] $\mb{x}$: position of first robot pose (estimate, in world coordinates), arbitrary, $2 \times 1$
\item[] $\mb{u}$: orientation vector of first robot pose (estimate, in world coordinates), arbitrary, $2 \times 1$
\item[] $\mb{U}$: orientation matrix of first robot pose (estimate), square, $2 \times 2$
\item[] $\mb{p}$: first robot pose (estimate, in world coordinates), arbitrary, $4 \times 1$
\item[] $\mb{x}'$: position of second robot pose (estimate, in world coordinates), arbitrary, $2 \times 1$
\item[] $\mb{u}'$: orientation vector of second robot pose (estimate, in world coordinates), arbitrary, $2 \times 1$
\item[] $\mb{U}'$: orientation matrix of second robot pose (estimate), square, $2 \times 2$
\item[] $\mb{p}'$: second robot pose (estimate, in world coordinates), arbitrary, $4 \times 1$
\end{description}
\textbf{Cost Functions}
\begin{description}
\setlength{\itemsep}{0pt}
\item[] $e$: distance odometry cost function, scalar, $1 \times 1$
\item[] $f$: translatory odometry cost function, scalar, $1 \times 1$
\item[] $g$: rotatory odometry cost function 1, scalar, $1 \times 1$
\item[] $\bar{g}$: rotatory odometry cost function 2, scalar, $1 \times 1$
\item[] $h$: home-vector cost function 1, scalar, $1 \times 1$
\item[] $\bar{h}$: home-vector cost function 2, scalar, $1 \times 1$
\item[] $c$: compass cost function 1, scalar, $1 \times 1$
\item[] $\bar{c}$: compass cost function 2, scalar, $1 \times 1$
\item[] $\gamma$: weight factor for orientation cost terms, scalar, $1 \times 1$
\item[] ${\lambda}_{i}$: Lagrange multipliers, scalar, $1 \times 1$
\item[] $\bs{\lambda}$: vector of Lagrange multipliers, arbitrary, $(N - 1) \times 1$
\item[] $w$: Lagrange constraint term on orientation vectors, scalar, $1 \times 1$
\item[] $l$: inner constraint term on orientation vectors, scalar, $1 \times 1$
\item[] $L$: Lagrangian, scalar, $1 \times 1$
\item[] $F$: sum of cost functions in Lagrangian, scalar, $1 \times 1$
\item[] $s$: generic rotational cost function 1, scalar, $1 \times 1$
\item[] $\bar{s}$: generic rotational cost function 2, scalar, $1 \times 1$
\item[] $\varphi$: angle, scalar, $1 \times 1$
\item[] $\bs{\Phi}$: orientation matrix, orthogonal, $2 \times 2$
\end{description}
\textbf{Newton Method}
\begin{description}
\setlength{\itemsep}{0pt}
\item[] $\mb{H}$: total Hessian (without first pose), symmetric, $(5 N - 5) \times (5 N - 5)$
\item[] $\mb{g}$: total gradient vector (without first pose), arbitrary, $(5 N - 5) \times 1$
\item[] $\bs{\Delta}\mb{s}$: state change vector (without first pose), arbitrary, $(5 N - 5) \times 1$
\item[] $\mb{R}$: regularization matrix, diagonal, $(5 N - 5) \times (5 N - 5)$
\item[] $\eta_W$: Levenberg-Marquardt factor (for core of bordered Hessian), scalar, $1 \times 1$
\item[] $\eta_A$: Levenberg-Marquardt factor (for border of bordered Hessian), scalar, $1 \times 1$
\end{description}
\textbf{Other}
\begin{description}
\setlength{\itemsep}{0pt}
\item[] $\mathfrak{d}$: Kronecker delta, scalar, $1 \times 1$
\item[] $\mb{M}$: transformation matrix, symmetric, $2 \times 2$
\item[] $\mb{N}$: transformation matrix, symmetric, $2 \times 2$
\item[] $\bs{\Omega}$: operator which turns orientation vector into orientation matrix, square, $2 \times 2$
\item[] $\overline{\bs{\Omega}}$: operator with negated second column w.r.t. $\bs{\Omega}$, symmetric, $2 \times 2$
\item[] $\mb{x}$: general vector, arbitrary, $2 \times 1$
\item[] $\mb{y}$: general vector, arbitrary, $2 \times 1$
\item[] $\mb{z}$: general vector, arbitrary, $2 \times 1$
\item[] $\mb{v}$: general vector, arbitrary, $2 \times 1$
\item[] $t_1$: configuration parameter, scalar, $1 \times 1$
\item[] $\bar{t}_1$: configuration parameter, scalar, $1 \times 1$
\end{description}

\section{Orientation Vectors}\label{sec_orvec}

%
In Lagrange-Newton SAM, angles are described by unit ``orientation
vectors''. In this way, the problem space remains Euclidean. Unit-length
constraints on the orientation vectors are handled by constraint terms
in the Lagrangian. To give an example of an orientation vector: If the
orientation of the first robot pose is described by an angle $\Theta$,
the orientation vector $\mb{u}$ is
\begin{equation}
    \mb{u} = \begin{pmatrix}u_{1}\\u_{2}\end{pmatrix} =
    \begin{pmatrix}
        \cos \Theta\\
        \sin \Theta
    \end{pmatrix}.
\end{equation}
Note that there are constant unit vectors (e.g. those relating to
measurements) and variable vectors which should approach unit length by
the constraints applied in the Lagrange-Newton descent ($\mb{u}$ is
actually an example of the latter class: it is only initially computed
as a unit vector from the angle $\Theta$ and then updated in the
Lagrange-Newton descent).

To form a Cartesian right-handed coordinate system where $\mb{u}$ is the
first axis, the vector perpendicular to $\mb{u}$ can be obtained from
\begin{align}
& \begin{pmatrix}- u_{2}\\u_{1}\end{pmatrix}.
\end{align}
The orthogonal (or approximately orthogonal) ``orientation matrix''
describing the coordinate system formed by the two vectors is expressed
by the operator $\bs{\Omega}$ applied to an orientation vector:
\begin{align}
\mb{U} & = \bs{\Omega}\mleft( \mb{u} \mright) = \begin{pmatrix}u_{1} & - u_{2}\\u_{2} & u_{1}\end{pmatrix}.
\end{align}
To add angles, one of the angles is turned into an (approximately)
orthogonal rotation matrix, namely the corresponding orientation matrix.
For example, if we want to add the odometry orientation change $\varphi$
(described by orientation vector $\mb{q}$ and the orientation matrix
$\mb{Q}$) to the orientation angle $\Theta$ (described by orientation
vector $\mb{u}$), we can write $\mb{Q} \mb{u} $.

To give a numerical example: for $\varphi = 30^\circ$, $\Theta =
60^\circ$ we get
\begin{align}
\mb{q}
&=
\begin{pmatrix}\frac{1}{2}\, \sqrt{3} \\\frac{1}{2}\,\end{pmatrix}\\
\mb{Q} & = \bs{\Omega}\mleft( \mb{q} \mright) = \begin{pmatrix}\frac{1}{2}\, \sqrt{3} & - \frac{1}{2}\,\\\frac{1}{2}\, &
\frac{1}{2}\, \sqrt{3} \end{pmatrix}\\
\mb{u}
&=
\begin{pmatrix}\frac{1}{2}\,\\\frac{1}{2}\, \sqrt{3} \end{pmatrix}\\
\mb{Q} \mb{u} & = \begin{pmatrix}\frac{1}{2}\, \sqrt{3} \frac{1}{2}\, + \mleft( -
\frac{1}{2}\, \frac{1}{2}\, \sqrt{3} \mright)\\\frac{1}{2}\,
\frac{1}{2}\, + \frac{1}{2}\, \sqrt{3} \frac{1}{2}\, \sqrt{3}
\end{pmatrix} = \begin{pmatrix}\frac{1}{4}\, \sqrt{3} + \mleft( - \frac{1}{4}\, \sqrt{3}
\mright)\\\frac{1}{4}\, + \frac{1}{4}\, \sqrt{3} \sqrt{3} \end{pmatrix} = \begin{pmatrix}0\\1\end{pmatrix},
\end{align}
which is the orientation vector for the angle sum $\varphi + \Theta =
90^\circ$.

In the derivatives, a second operator appears which is labeled
$\overline{\bs{\Omega}}$ (see below). It differs from $\bs{\Omega}$ in
that the second column is multiplied by $-1$. Applied to $\mb{u}$ it
would give
\begin{align}
\overline{\bs{\Omega}}\mleft( \mb{u} \mright) & = \begin{pmatrix}u_{1} & u_{2}\\u_{2} & - u_{1}\end{pmatrix}.
\end{align}
For transformations, we introduce the symmetric matrices $\mb{M}$ and
$\mb{N}$
\begin{align}
\mb{M}
&=
\begin{pmatrix}1 & 0\\0 & - 1\end{pmatrix}\\
\mb{N}
&=
\begin{pmatrix}0 & 1\\1 & 0\end{pmatrix}
\end{align}
such that we can write
\begin{align}
\mb{U} & = \bs{\Omega}\mleft( \mb{u} \mright) = \begin{pmatrix}\mb{u}^T \mb{M} \\\mb{u}^T \mb{N} \end{pmatrix}\\
\mb{U}^T & = \mleft( \bs{\Omega}\mleft( \mb{u} \mright) \mright)^T = \begin{pmatrix}\mb{M} \mb{u} & \mb{N} \mb{u} \end{pmatrix}.
\end{align}
We generally see that, for any two-dimensional vectors $\mb{x}$,
$\mb{y}$, and $\mb{z}$,
\begin{align}
\bs{\Omega}\mleft( \mb{z} \mright) & = \begin{pmatrix}\mb{z}^T \mb{M} \\\mb{z}^T \mb{N} \end{pmatrix} = \begin{pmatrix}z_{1} & - z_{2}\\z_{2} & z_{1}\end{pmatrix}\\
\mleft( \bs{\Omega}\mleft( \mb{z} \mright) \mright)^T & = \begin{pmatrix}\mb{M} \mb{z} & \mb{N} \mb{z} \end{pmatrix}\\
\overline{\bs{\Omega}}\mleft( \mb{z} \mright) & = \bs{\Omega}\mleft( \mb{z} \mright) \mb{M}\\
\overline{\bs{\Omega}}\mleft( \mb{z} \mright) & = \mb{M} z_{1} + \mb{N} z_{2} = \begin{pmatrix}1 & 0\\0 & - 1\end{pmatrix} z_{1} + \begin{pmatrix}0 &
1\\1 & 0\end{pmatrix} z_{2} = \begin{pmatrix}z_{1} & z_{2}\\z_{2} & - z_{1}\end{pmatrix}\\
\bs{\Omega}\mleft( \mb{x} + \mb{y} \mright) & = \begin{pmatrix}x_{1} + y_{1} & - \mleft( x_{2} + y_{2} \mright)\\x_{2} +
y_{2} & x_{1} + y_{1}\end{pmatrix} = \begin{pmatrix}x_{1} & - x_{2}\\x_{2} & x_{1}\end{pmatrix} +
\begin{pmatrix}y_{1} & - y_{2}\\y_{2} & y_{1}\end{pmatrix} = \bs{\Omega}\mleft( \mb{x} \mright) + \bs{\Omega}\mleft( \mb{y} \mright).
\end{align}
The derivative of a product of a transposed orientation matrix with a
vector $\mb{x}$ (i.e. a projection) introduces the operator
$\overline{\bs{\Omega}}$:
\begin{align}
\label{eq_Omega_deriv_OmegaBar}
\frac{\partial}{\partial \mb{z}}\,\mleft( \mleft[ \bs{\Omega}\mleft(
\mb{z} \mright) \mright]^T \mb{x} \mright) & = \frac{\partial}{\partial \mb{z}}\,\mleft( \begin{pmatrix}\mb{M} \mb{z} &
\mb{N} \mb{z} \end{pmatrix} \mb{x} \mright) = \frac{\partial}{\partial \mb{z}}\,\mleft( \mb{M} \mb{z} x_{1} + \mb{N}
\mb{z} x_{2} \mright) = \mb{M} x_{1} + \mb{N} x_{2} = \overline{\bs{\Omega}}\mleft( \mb{x} \mright).
\end{align}
The derivative of a product of an orientation matrix with a vector
$\mb{x}$ (i.e. a rotation) leads to:
\begin{align}
&\phantom{{}={}} \frac{\partial}{\partial \mb{z}}\,\mleft( \bs{\Omega}\mleft( \mb{z}
\mright) \mb{x} \mright) \nonumber \\
& = \frac{\partial}{\partial \mb{z}}\,\mleft( \begin{pmatrix}\mb{z}^T \mb{M}
\\\mb{z}^T \mb{N} \end{pmatrix} \mb{x} \mright)\\
& = \frac{\partial}{\partial \mb{z}}\,\mleft( \begin{pmatrix}\mb{z}^T \mb{M}
\mb{x} \\\mb{z}^T \mb{N} \mb{x} \end{pmatrix} \mright)\\
& = \frac{\partial}{\partial \mb{z}}\,\mleft( \begin{pmatrix}\mb{x}^T \mb{M}
\mb{z} \\\mb{x}^T \mb{N} \mb{z} \end{pmatrix} \mright)\\
& = \begin{pmatrix}\mb{x}^T \mb{M} \\\mb{x}^T \mb{N} \end{pmatrix}\\
& = \bs{\Omega}\mleft( \mb{x} \mright).
\end{align}

\section{Cost Functions}\label{sec_cost}

\begin{figure}[t]
\begin{center}
\includegraphics[width=\textwidth]{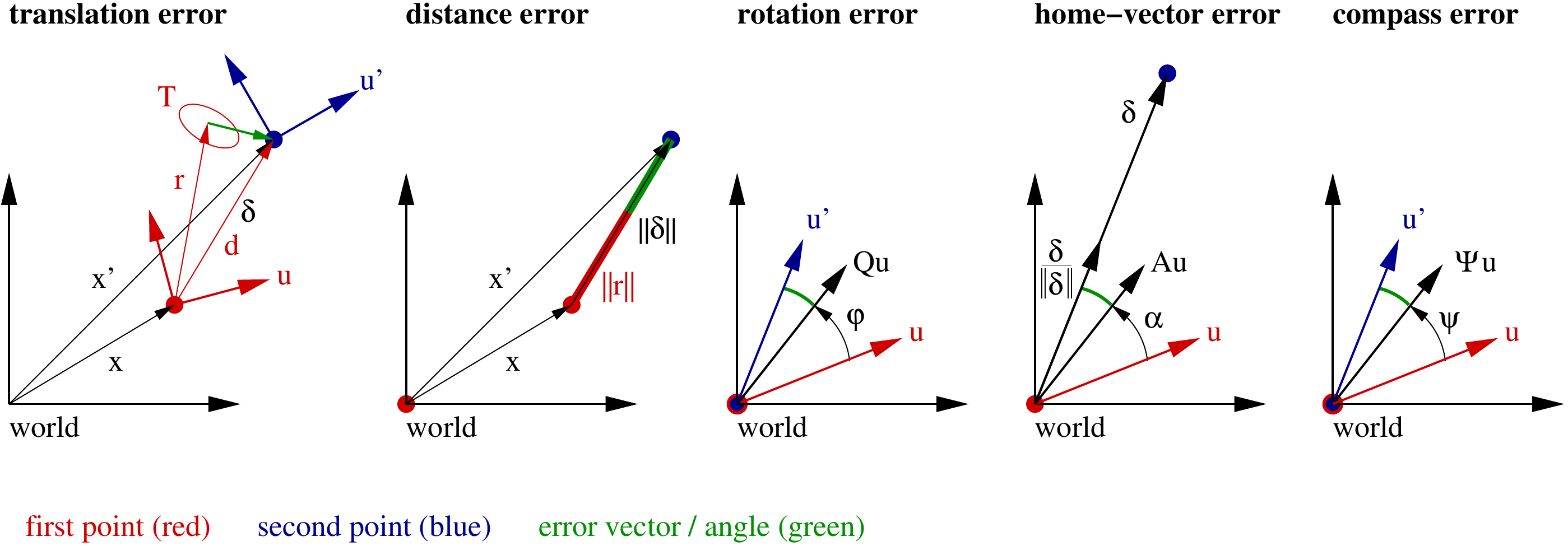}

\caption{Cost functions. Each diagram shows the computation of the
cost term for a pair of poses in the plane (red and blue dots and
orientation vectors). Green distances and green angles visualize
errors. The uncertainty is only visualized for the translation
error (error ellipse); for the other cost functions, the
one-dimensional translatory and angular uncertainties are not
shown.}

\label{fig_costfct}
\end{center}
\end{figure}

%
Figure \ref{fig_costfct} provides a diagram explaining the derivation of
each of the five cost functions.

\subsection{Pose Vectors}\label{sec_cost_pose}

%
The following equations use the pose vectors $\mb{p}$ (first pose) and
$\mb{p}'$ (second pose), which are defined as:
\begin{align}
\mb{p}
&=
\begin{pmatrix}\mb{x}\\\mb{u}\end{pmatrix}\\
\mb{p}'
&=
\begin{pmatrix}\mb{x}'\\\mb{u}'\end{pmatrix},
\end{align}
where $\mb{x}$ and $\mb{x}'$ are 2D position vectors and $\mb{u}$ and
$\mb{u}'$ are 2D orientation vectors.

These pose vectors are estimates (in world coordinates) refined by the
Lagrange-Newton method. Therefore $\mb{u}$ and $\mb{u}'$ can deviate
from unit-length in the course of the computation (but should approach
unit length near the solution).

For odometry measurements, first and second pose vectors come from
subsequent time steps of the robot's trajectory. For homing
measurements, the first pose vector relates to the current pose, the
second pose vector to an earlier pose; the view captured at the earlier
pose is used as a landmark.

\subsection{Cost Functions and Constraint}\label{sec_cost_constraint}

%
An issue to address in the design of the cost functions is how they are
affected by the unit-length constraint on the orientation vectors they
are depending on. The {\em weakest requirement} is that the cost
functions show the desired effect at least if the constraint is
fulfilled. This may allow for negative costs which is at least unusual
and could possibly lead to problems in the Lagrange-Newton descent (e.g.
getting stuck in compromise solutions). An {\em intermediate
requirement} is therefore that the cost functions are in addition
non-negative. The {\em strongest requirement} is that the cost functions
are both non-negative and invariant of constraint violations. The
complexity of cost functions and their derivatives grows with the
strength of the requirement.

The question is whether and how the three requirement levels affect the
performance of Lagrange-Newton SAM. Below we explore the alternatives
for all rotational cost functions by defining three different variants
bundled in two forms. This concerns the rotation error (odometry) and
the home-vector and compass error (homing). The translation error
(odometry) fulfills the intermediate requirement, but no version with
the strongest requirement has been explored yet. The distance error
(odometry) does not include orientation vectors and is therefore not
affected.

\subsection{Generic Rotational Cost Function}\label{sec_cost_genrot}

%
Below we define two purely rotational cost functions (rotation error and
compass error) which only depend on $\mb{u}$ and $\mb{u}'$. Both have
the same form and can therefore be treated in the same way. (Note that
the home-vector error is also a rotational cost function, but has a
special form and will therefore be treated separately.) Let $\bs{\Phi}$
be an orthogonal orientation matrix (corresponding to an angle
$\varphi$). We explore three alternative variants of rotational cost
functions:
\begin{align}
s\mleft( \bs{\Phi},\mb{p},\mb{p}' \mright)
&=
1 - \mleft( \bs{\Phi} \mb{u} \mright)^T \mb{u}'\\
s\mleft( \bs{\Phi},\mb{p},\mb{p}' \mright)
&=
\mleft\| \mb{u} \mright\| \mleft\| \mb{u}' \mright\| - \mleft( \bs{\Phi}
\mb{u} \mright)^T \mb{u}'\\
\label{eq_second_form}
\bar{s}\mleft( \bs{\Phi},\mb{p},\mb{p}' \mright)
&=
1 - \mleft( \bs{\Phi} {\,\frac{\mb{u}}{\mleft\| \mb{u} \mright\|}\,}
\mright)^T {\,\frac{\mb{u}'}{\mleft\| \mb{u}' \mright\|}\,}.
\end{align}
The term $\bs{\Phi} \mb{u} $ expresses a rotation of the first
orientation vector $\mb{u}$ by the angle $\varphi$ expressed by the
orthogonal orientation matrix $\bs{\Phi}$. The rotated orientation
vector $\bs{\Phi} \mb{u} $ is matched to the second orientation vector
$\mb{u}'$. (In the third variant, $\mb{u}$ and $\mb{u}'$ are
normalized.) The generic rotational cost functions do not include
uncertainty terms; these are introduced later for the specific cost
functions.

In the first variant above (weakest requirement, see above), we
determine a scalar product between $\bs{\Phi} \mb{u} $ and $\mb{u}'$ and
subtract it from $1$ to obtain a distance measure. If $\mb{u}$ and
$\mb{u}'$ are unit vectors, the last factor corresponds to $1 - \cos
\beta$ where $\beta$ is the angle between $\bs{\Phi} \mb{u} $ and
$\mb{u}'$. The Taylor expansion of $\cos \beta$ for small $\beta$ gives
$\cos \beta \approx 1 - \frac{1}{2} \beta ^2$, so for small $\beta$ we
have a quadratic error as for the translation error (see below). If
$\mb{u}$ or $\mb{u}'$ deviate from unit length, the costs can become
negative in the first variant.

To avoid negative costs while keeping the cost function simple, we
introduced the second variant (intermediate requirement) which differs
in the first summand. Non-negativity is ensured by exploiting the
Cauchy-Schwarz inequality (note that $\bs{\Phi}$ is orthogonal and the
norm is invariant to multiplication by an orthogonal matrix). This
variant is still affected by constraint violations.

The third variant normalizes the orientation vectors $\mb{u}$ and
$\mb{u}'$ to unit length such that the costs are always non-negative and
the cost function is invariant of constraint violations (strongest
requirement).

The first two variants are similar, so we fuse them using configuration
parameters $t_1$ and $\bar{t}_1 = 1 - t_1$ with $t_1 \in \{0, 1\}$:
\begin{align}
\label{eq_first_form}
s\mleft( \bs{\Phi},\mb{p},\mb{p}' \mright)
&=
t_1 + \bar{t}_1 \mleft\| \mb{u} \mright\| \mleft\| \mb{u}' \mright\| -
\mleft( \bs{\Phi} \mb{u} \mright)^T \mb{u}'.
\end{align}
This form is later referred to as ``first form'', while equation
\eqref{eq_second_form} is referred to as ``second form''.

In all rotational cost functions introduced below, the same factor
$\gamma$ is used as a weight factor (while translatory cost functions
implicitly have unit weight). Even though rotatory and translatory cost
functions are similar for small deviations, rotational cost terms are
still different in character from translatory terms for larger
deviations, as they are restricted in range (at least on the constraint)
and cyclic.

\subsection{Odometry Measurement}\label{sec_cost_odo}

%
The odometry measurement between the first and second robot pose is
described by vector $\mb{r}$ (in robot coordinates of the first robot
pose) and orientation change vector $\mb{q}$ (or the corresponding
orientation matrix $\mb{Q}$).

The uncertainty of the pose is described by the covariance matrix
$\mb{C}$ (in robot coordinates of the first pose):
\begin{align}
\mb{C}
&=
\begin{pmatrix}\mb{T} & \mb{b}\\\mb{b}^T & \sigma^{2}\end{pmatrix}
\end{align}
We ignore the cross-covariance terms $\mb{b}$ and $\mb{b}^T$ and
approximate\footnote{The effects of this approximation on the inverse of
$\mb{C}$ should be explored.}
\begin{align}
\mb{C}
&\approx
\begin{pmatrix}\mb{T} & \mb{0}_{2}\\\mb{0}_{2}^T &
\sigma^{2}\end{pmatrix}.
\end{align}
Here, the covariance matrix $\mb{T}$ describes the translatory
uncertainty, the standard deviation $\sigma$ the angular uncertainty. In
the following, we treat the translation error and the rotation error
separately. In addition, we define a distance error which depends on the
difference in measured and estimated distance between first and second
position.

\subsubsection{Translation Error}\label{sec_cost_trans}

%
The translation error is defined as a Mahalanobis distance:
\begin{align}
\label{eq_trans_error}
f\mleft( \mb{p},\mb{p}' \mright)
&=
\frac{1}{2}\, \mleft( \mb{d} - \mb{r} \mright)^T \mb{T}^{-1} \mleft(
\mb{d} - \mb{r} \mright)
\end{align}
where $\mb{d} = \mb{d}\mleft( \mb{x},\mb{u},\mb{x}' \mright)$ is the
estimated position change in robot coordinates of the first pose,
obtained from the estimated position change in world coordinates
$\bs{\delta} = \mb{x}' - \mb{x}$:
\begin{align}
\mb{d} & = \mb{U}^T \bs{\delta} = \mb{U}^T \mleft( \mb{x}' - \mb{x} \mright) = \begin{pmatrix}u_{1} & - u_{2}\\u_{2} & u_{1}\end{pmatrix}^T \mleft(
\mb{x}' - \mb{x} \mright).
\end{align}

\subsubsection{Distance Error}\label{sec_cost_dist}

%
The distance error is defined with measured distance $\varrho = \mleft\|
\mb{r} \mright\|$ and estimated distance $\mleft\| \bs{\delta}
\mright\|$ as:
\begin{align}
\label{eq_dist_error}
e\mleft( \mb{p},\mb{p}' \mright) & = \frac{1}{2}\, {\,\frac{1}{\sigma_e}\,} \mleft( \mleft\| \mb{x}' - \mb{x}
\mright\| - \varrho \mright)^2 = \frac{1}{2}\, {\,\frac{1}{\sigma_e}\,} \mleft( \mleft\| \bs{\delta}
\mright\| - \varrho \mright)^2.
\end{align}
The distance error is introduced as a second example of a translatory
cost function and for tests of the implementation, but is not used in
the experiments described below.

\subsubsection{Rotation Error}\label{sec_const_rot}

%
The rotation error is defined in two forms by
\begin{align}
\label{eq_rot_error}
g\mleft( \mb{p},\mb{p}' \mright) & = \gamma {\,\frac{1}{\sigma^{2}}\,} s\mleft( \mb{Q},\mb{p},\mb{p}'
\mright) = \gamma {\,\frac{1}{\sigma^{2}}\,} \mleft( t_1 + \bar{t}_1 \mleft\|
\mb{u} \mright\| \mleft\| \mb{u}' \mright\| - \mleft[ \mb{Q} \mb{u}
\mright]^T \mb{u}' \mright)\\
\label{eq_rot_error_norm}
\bar{g}\mleft( \mb{p},\mb{p}' \mright) & = \gamma {\,\frac{1}{\sigma^{2}}\,} \bar{s}\mleft( \mb{Q},\mb{p},\mb{p}'
\mright) = \gamma {\,\frac{1}{\sigma^{2}}\,} \mleft( 1 - \mleft[ \mb{Q}
{\,\frac{\mb{u}}{\mleft\| \mb{u} \mright\|}\,} \mright]^T
{\,\frac{\mb{u}'}{\mleft\| \mb{u}' \mright\|}\,} \mright).
\end{align}
It depends on the cosine of the angle between the orientation vector of
the first pose, rotated by the odometry rotation measurement (robot
coordinates), and the orientation vector of the second pose.

\subsection{Homing Measurement}\label{sec_cost_homing}

%
We assume that a homing measurement delivers two orientation vectors
with unit length: $\bs{\alpha}$ (or the corresponding orientation matrix
$\bs{A}$) points approximately from the first to the second pose (in
robot coordinates of the first pose), and $\bs{\psi}$ (or the
corresponding orientation matrix $\bs{\Psi}$) is a measurement of the
orientation difference between the second and the first pose.

The uncertainty of these two measurements is described by the standard
deviations $\sigma_h$ (home vector) and $\sigma_c$ (compass).

\subsubsection{Home-Vector Error}\label{sec_cost_homevector}

%
The home-vector error is defined in two forms by
\begin{align}
\label{eq_homevector_error}
h\mleft( \mb{p},\mb{p}' \mright) & = \gamma {\,\frac{1}{\sigma_h^{2}}\,} \mleft( t_1 + \bar{t}_1 \mleft\|
\mb{u} \mright\| - \mleft[ \bs{A} \mb{u} \mright]^T
{\,\frac{\bs{\delta}}{\mleft\| \bs{\delta} \mright\|}\,} \mright)\\
\label{eq_homevector_error_norm}
\bar{h}\mleft( \mb{p},\mb{p}' \mright) & = \gamma {\,\frac{1}{\sigma_h^{2}}\,} \mleft( 1 - \mleft[ \bs{A}
{\,\frac{\mb{u}}{\mleft\| \mb{u} \mright\|}\,} \mright]^T
{\,\frac{\bs{\delta}}{\mleft\| \bs{\delta} \mright\|}\,} \mright).
\end{align}
It depends on the cosine of the angle between the orientation vector of
the first pose, rotated by the home-vector angle measurement (robot
coordinates), and the unit vector pointing from the first to the second
pose. Note that this is a rotational measure (as the rotation error
above and the compass error below), but the second orientation vector is
determined from the normalized home vector and not from the orientation
vector of a pose.

\subsubsection{Compass Error}\label{sec_const_compass}

%
The compass error is defined in two forms by
\begin{align}
\label{eq_compass_error}
c\mleft( \mb{p},\mb{p}' \mright) & = \gamma {\,\frac{1}{\sigma_c^{2}}\,} s\mleft( \bs{\Psi},\mb{p},\mb{p}'
\mright) = \gamma {\,\frac{1}{\sigma_c^{2}}\,} \mleft( t_1 + \bar{t}_1 \mleft\|
\mb{u} \mright\| \mleft\| \mb{u}' \mright\| - \mleft[ \bs{\Psi} \mb{u}
\mright]^T \mb{u}' \mright)\\
\label{eq_compass_error_norm}
\bar{c}\mleft( \mb{p},\mb{p}' \mright) & = \gamma {\,\frac{1}{\sigma_c^{2}}\,} \bar{s}\mleft(
\bs{\Psi},\mb{p},\mb{p}' \mright) = \gamma {\,\frac{1}{\sigma_c^{2}}\,} \mleft( 1 - \mleft[ \bs{\Psi}
{\,\frac{\mb{u}}{\mleft\| \mb{u} \mright\|}\,} \mright]^T
{\,\frac{\mb{u}'}{\mleft\| \mb{u}' \mright\|}\,} \mright).
\end{align}
It depends on the cosine of the angle between the orientation vector of
the first pose, rotated by the compass measurement (robot coordinates),
and the orientation vector of the second pose.

\section{Derivatives}\label{sec_deriv}

%
In the following, we compute the first and second derivatives of all
cost functions. Zero derivatives are omitted. For some steps, we used
the Matrix Calculus tool at \url{www.matrixcalculus.org}
\citep[][]{nn_Laue18,nn_Laue20}, particularly the following derivatives
\begin{align}
\frac{\partial}{\partial \mb{x}}\,\mleft\| \mb{x} \mright\| & = {\,\frac{\mb{x}^T}{\mleft\| \mb{x} \mright\|}\,} = \mb{x}_0^T\\
\frac{\partial}{\partial \mb{x}}\,{\,\frac{\mb{x}}{\mleft\| \mb{x}
\mright\|}\,} & = {\,\frac{\mb{I}_{2}}{\mleft\| \mb{x} \mright\|}\,} - {\,\frac{\mb{x}
\mb{x}^T }{\mleft\| \mb{x} \mright\|^{3}}\,} = {\,\frac{1}{\mleft\| \mb{x} \mright\|}\,} \mleft( \mb{I}_{2} - \mb{x}_0
\mb{x}_0^T \mright)\\
\frac{\partial}{\partial \mb{x}}\,\mleft( \mleft\| \mb{x} \mright\|
\mb{v} \mright) & = {\,\frac{\mb{v} \mb{x}^T }{\mleft\| \mb{x} \mright\|}\,} = \mb{v} \mb{x}_0^T\\
\frac{\partial}{\partial \mb{x}}\,{\,\frac{\mb{v}}{\mleft\| \mb{x}
\mright\|}\,} & = - {\,\frac{\mb{v} \mb{x}^T }{\mleft\| \mb{x} \mright\|^{3}}\,} = - {\,\frac{\mb{v} \mb{x}_0^T }{\mleft\| \mb{x} \mright\|^{2}}\,}
\end{align}
\parbox{\textwidth}{\rule{\textwidth}{0.4pt}}
\begin{align}
&\phantom{{}={}} \frac{\partial}{\partial \mb{x}}\,{\,\frac{\mb{x} \mb{x}^T \mb{v}
}{\mleft\| \mb{x} \mright\|^{3}}\,} \nonumber \\
& = {\,\frac{\mleft( \mb{x}^T \mb{v} \mright) \mb{I}_{2} }{\mleft\| \mb{x}
\mright\|^{3}}\,} - 3 {\,\frac{\mleft( \mb{x}^T \mb{v} \mright) \mb{x}
\mb{x}^T }{\mleft\| \mb{x} \mright\|^{5}}\,} + {\,\frac{\mb{x} \mb{v}^T
}{\mleft\| \mb{x} \mright\|^{3}}\,}\\
& = {\,\frac{1}{\mleft\| \mb{x} \mright\|^{2}}\,} \mleft( \mleft[ \mb{x}_0^T
\mb{v} \mright] \mb{I}_{2} - 3 \mleft[ \mb{x}_0^T \mb{v} \mright]
\mb{x}_0 \mb{x}_0^T + \mb{x}_0 \mb{v}^T \mright)\\
& = {\,\frac{1}{\mleft\| \mb{x} \mright\|^{2}}\,} \mleft( \mleft[ \mb{x}_0^T
\mb{v} \mright] \mleft[ \mb{I}_{2} - 3 \mb{x}_0 \mb{x}_0^T \mright] +
\mb{x}_0 \mb{v}^T \mright)
\end{align}
\parbox{\textwidth}{\rule{\textwidth}{0.4pt}}
\begin{align}
&\phantom{{}={}} \frac{\partial}{\partial \mb{x}}\,\mleft( \mleft[
{\,\frac{\mb{I}_{2}}{\mleft\| \mb{x} \mright\|}\,} - {\,\frac{\mb{x}
\mb{x}^T }{\mleft\| \mb{x} \mright\|^{3}}\,} \mright] \mb{v} \mright) \nonumber \\
& = - {\,\frac{1}{\mleft\| \mb{x} \mright\|^{2}}\,} \mleft( {\,\frac{\mb{v}
\mb{x}^T }{\mleft\| \mb{x} \mright\|}\,} + {\,\frac{\mleft[ \mb{x}^T
\mb{v} \mright] \mb{I}_{2} }{\mleft\| \mb{x} \mright\|}\,} - {\,\frac{3
\mleft[ \mb{x}^T \mb{v} \mright] \mb{x} \mb{x}^T }{\mleft\| \mb{x}
\mright\|^{3}}\,} + {\,\frac{\mb{x} \mb{v}^T }{\mleft\| \mb{x}
\mright\|}\,} \mright)\\
& = - {\,\frac{1}{\mleft\| \mb{x} \mright\|^{2}}\,} \mleft( \mb{v}
\mb{x}_0^T + \mleft[ \mb{x}_0^T \mb{v} \mright] \mb{I}_{2} - 3 \mleft[
\mb{x}_0^T \mb{v} \mright] \mb{x}_0 \mb{x}_0^T + \mb{x}_0 \mb{v}^T
\mright)\\
& = - {\,\frac{1}{\mleft\| \mb{x} \mright\|^{2}}\,} \mleft( \mb{v}
\mb{x}_0^T + \mleft[ \mb{x}_0^T \mb{v} \mright] \mleft[ \mb{I}_{2} - 3
\mb{x}_0 \mb{x}_0^T \mright] + \mb{x}_0 \mb{v}^T \mright)
\end{align}
where
\begin{align}
\mb{x}_0
&=
{\,\frac{\mb{x}}{\mleft\| \mb{x} \mright\|}\,}.
\end{align}

\subsection{First Derivatives}\label{sec_first}

%
In the following, we apply the chain rule in vector form \citep[see
e.g.][p.129]{nn_Deisenroth20}. Dimensions are $n$: $1 \leq n$; $m$: $1
\leq m$, variables and functions are $\mb{x}$: arbitrary, $n \times 1$;
$\mb{z}$: arbitrary, $m \times 1$; $\mb{f}$: arbitrary, $m \times 1$;
$g$: scalar, $1 \times 1$; $y$: scalar, $1 \times 1$:
\begin{align}
\mb{z}
&=
\mb{f}\mleft( \mb{x} \mright)\\
y
&=
g\mleft( \mb{z} \mright)\\
\frac{\partial}{\partial \mb{x}}\,g
&=
\mleft( \frac{\partial}{\partial \mb{z}}\,g \mright) \mleft(
\frac{\partial}{\partial \mb{x}}\,\mb{f} \mright).
\end{align}
Note that the first factor (a gradient) is a row vector, and the second
factor is a Jacobian matrix; their product is a row vector.
The chain rule can also be applied for multi-step dependencies between
vectors, e.g. in
\begin{align}
\frac{\partial}{\partial \mb{x}}\,h\mleft( \mb{p},\mb{p}^T \mright) & = \mleft( \frac{\partial}{\partial \bs{\delta}}\,h\mleft( \mb{p},\mb{p}^T
\mright) \mright) \mleft( \frac{\partial}{\partial \mb{x}}\,\bs{\delta}
\mright) = \mleft( \frac{\partial}{\partial \bs{\delta}_0}\,h\mleft(
\mb{p},\mb{p}^T \mright) \mright) \mleft( \frac{\partial}{\partial
\bs{\delta}}\,\bs{\delta}_0 \mright) \mleft( \frac{\partial}{\partial
\mb{x}}\,\bs{\delta} \mright).
\end{align}

\subsection{First Derivatives of Generic Rotational Cost Function}\label{sec_first_genrot}

%
We define
\begin{align}
\mb{u}_0
&=
{\,\frac{\mb{u}}{\mleft\| \mb{u} \mright\|}\,}\\
{{} {\mb{u}'_0}}
&=
{\,\frac{\mb{u}'}{\mleft\| \mb{u}' \mright\|}\,}
\end{align}
and get for the first form
\begin{align}
&\phantom{{}={}} \frac{\partial}{\partial \mb{u}}\,s\mleft( \bs{\Phi},\mb{p},\mb{p}'
\mright) \nonumber \\
& = \frac{\partial}{\partial \mb{u}}\,\mleft( t_1 + \bar{t}_1 \mleft\|
\mb{u} \mright\| \mleft\| \mb{u}' \mright\| - \mleft[ \bs{\Phi} \mb{u}
\mright]^T \mb{u}' \mright)\\
& = \frac{\partial}{\partial \mb{u}}\,\mleft( t_1 + \bar{t}_1 \mleft\|
\mb{u} \mright\| \mleft\| \mb{u}' \mright\| - \mb{u}'^T \bs{\Phi} \mb{u}
\mright)\\
& = \bar{t}_1 \mleft\| \mb{u}' \mright\| {\,\frac{\mb{u}^T}{\mleft\| \mb{u}
\mright\|}\,} - \mb{u}'^T \bs{\Phi}\\
&\label{eq_first_genrot_u}
 = \bar{t}_1 \mleft\| \mb{u}' \mright\| \mb{u}_0^T - \mb{u}'^T \bs{\Phi}
\end{align}
\parbox{\textwidth}{\rule{\textwidth}{0.4pt}}
\begin{align}
&\phantom{{}={}} \frac{\partial}{\partial \mb{u}'}\,s\mleft( \bs{\Phi},\mb{p},\mb{p}'
\mright) \nonumber \\
& = \frac{\partial}{\partial \mb{u}'}\,\mleft( t_1 + \bar{t}_1 \mleft\|
\mb{u} \mright\| \mleft\| \mb{u}' \mright\| - \mleft[ \bs{\Phi} \mb{u}
\mright]^T \mb{u}' \mright)\\
& = \frac{\partial}{\partial \mb{u}'}\,\mleft( t_1 + \bar{t}_1 \mleft\|
\mb{u} \mright\| \mleft\| \mb{u}' \mright\| - \mb{u}^T \bs{\Phi}^T
\mb{u}' \mright)\\
& = \bar{t}_1 \mleft\| \mb{u} \mright\| {\,\frac{\mb{u}'^T}{\mleft\| \mb{u}'
\mright\|}\,} - \mb{u}^T \bs{\Phi}^T\\
&\label{eq_first_genrot_us}
 = \bar{t}_1 \mleft\| \mb{u} \mright\| {{} {\mb{u}'_0}}^T - \mb{u}^T
\bs{\Phi}^T.
\end{align}
For the second form, we get
\begin{align}
&\phantom{{}={}} \frac{\partial}{\partial \mb{u}}\,\bar{s}\mleft(
\bs{\Phi},\mb{p},\mb{p}' \mright) \nonumber \\
& = \frac{\partial}{\partial \mb{u}}\,\mleft( 1 - \mleft[ \bs{\Phi}
{\,\frac{\mb{u}}{\mleft\| \mb{u} \mright\|}\,} \mright]^T
{\,\frac{\mb{u}'}{\mleft\| \mb{u}' \mright\|}\,} \mright)\\
& = \frac{\partial}{\partial \mb{u}}\,\mleft( 1 -
{\,\frac{\mb{u}'^T}{\mleft\| \mb{u}' \mright\|}\,} \bs{\Phi}
{\,\frac{\mb{u}}{\mleft\| \mb{u} \mright\|}\,} \mright)\\
& = - {\,\frac{\mb{u}'^T}{\mleft\| \mb{u}' \mright\|}\,} \bs{\Phi} \mleft(
{\,\frac{\mb{I}_{2}}{\mleft\| \mb{u} \mright\|}\,} - {\,\frac{\mb{u}
\mb{u}^T }{\mleft\| \mb{u} \mright\|^{3}}\,} \mright)\\
&\label{eq_first_genrot_u_norm}
 = - {{} {\mb{u}'_0}}^T \bs{\Phi} {\,\frac{1}{\mleft\| \mb{u} \mright\|}\,}
\mleft( \mb{I}_{2} - \mb{u}_0 \mb{u}_0^T \mright)
\end{align}
\parbox{\textwidth}{\rule{\textwidth}{0.4pt}}
\begin{align}
&\phantom{{}={}} \frac{\partial}{\partial \mb{u}'}\,\bar{s}\mleft(
\bs{\Phi},\mb{p},\mb{p}' \mright) \nonumber \\
& = \frac{\partial}{\partial \mb{u}'}\,\mleft( 1 - \mleft[ \bs{\Phi}
{\,\frac{\mb{u}}{\mleft\| \mb{u} \mright\|}\,} \mright]^T
{\,\frac{\mb{u}'}{\mleft\| \mb{u}' \mright\|}\,} \mright)\\
& = \frac{\partial}{\partial \mb{u}'}\,\mleft( 1 -
{\,\frac{\mb{u}^T}{\mleft\| \mb{u} \mright\|}\,} \bs{\Phi}^T
{\,\frac{\mb{u}'}{\mleft\| \mb{u}' \mright\|}\,} \mright)\\
& = - {\,\frac{\mb{u}^T}{\mleft\| \mb{u} \mright\|}\,} \bs{\Phi}^T \mleft(
{\,\frac{\mb{I}_{2}}{\mleft\| \mb{u}' \mright\|}\,} - {\,\frac{\mb{u}'
\mb{u}'^T }{\mleft\| \mb{u}' \mright\|^{3}}\,} \mright)\\
&\label{eq_first_genrot_us_norm}
 = - \mb{u}_0^T \bs{\Phi}^T {\,\frac{1}{\mleft\| \mb{u}' \mright\|}\,}
\mleft( \mb{I}_{2} - {{} {\mb{u}'_0}} {{} {\mb{u}'_0}}^T \mright).
\end{align}

\subsection{First Derivatives of Translation Error}\label{sec_first_trans}

%
We compute the first derivatives of the translation error
\eqref{eq_trans_error}:
\begin{align}
&\phantom{{}={}} \frac{\partial}{\partial \mb{x}}\,f\mleft( \mb{p},\mb{p}' \mright) \nonumber \\
& = \mleft( \frac{\partial}{\partial \mb{d}}\,f\mleft( \mb{p},\mb{p}'
\mright) \mright) \mleft( \frac{\partial}{\partial
\mb{x}}\,\mb{d}\mleft( \mb{x},\mb{u},\mb{x}' \mright) \mright)\\
& = \mleft( \mb{d}\mleft( \mb{x},\mb{u},\mb{x}' \mright) - \mb{r} \mright)^T
\mb{T}^{-1} \mleft( \frac{\partial}{\partial \mb{x}}\,\mleft[ \mb{U}^T
\mleft\{ \mb{x}' - \mb{x} \mright\} \mright] \mright)\\
& = - \mleft( \mb{d}\mleft( \mb{x},\mb{u},\mb{x}' \mright) - \mb{r}
\mright)^T \mb{T}^{-1} \mb{U}^T
\end{align}
\parbox{\textwidth}{\rule{\textwidth}{0.4pt}}
\begin{align}
&\phantom{{}={}} \frac{\partial}{\partial \mb{x}'}\,f\mleft( \mb{p},\mb{p}' \mright) \nonumber \\
& = \mleft( \frac{\partial}{\partial \mb{d}}\,f\mleft( \mb{p},\mb{p}'
\mright) \mright) \mleft( \frac{\partial}{\partial
\mb{x}'}\,\mb{d}\mleft( \mb{x},\mb{u},\mb{x}' \mright) \mright)\\
& = \mleft( \mb{d}\mleft( \mb{x},\mb{u},\mb{x}' \mright) - \mb{r} \mright)^T
\mb{T}^{-1} \mleft( \frac{\partial}{\partial \mb{x}'}\,\mleft[ \mb{U}^T
\mleft\{ \mb{x}' - \mb{x} \mright\} \mright] \mright)\\
& = \mleft( \mb{d}\mleft( \mb{x},\mb{u},\mb{x}' \mright) - \mb{r} \mright)^T
\mb{T}^{-1} \mb{U}^T.
\end{align}
\parbox{\textwidth}{\rule{\textwidth}{0.4pt}}
With the following intermediate result (see also equation
\eqref{eq_Omega_deriv_OmegaBar})
\begin{align}
&\phantom{{}={}} \frac{\partial}{\partial \mb{u}}\,\mb{d}\mleft( \mb{x},\mb{u},\mb{x}'
\mright) \nonumber \\
& = \frac{\partial}{\partial \mb{u}}\,\mleft( \mb{U}^T \bs{\delta} \mright)\\
& = \frac{\partial}{\partial \mb{u}}\,\mleft( \begin{pmatrix}u_{1} & -
u_{2}\\u_{2} & u_{1}\end{pmatrix}^T \bs{\delta} \mright)\\
& = \frac{\partial}{\partial \mb{u}}\,\mleft( \begin{pmatrix}u_{1} &
u_{2}\\- u_{2} & u_{1}\end{pmatrix}
\begin{pmatrix}\delta_{1}\\\delta_{2}\end{pmatrix} \mright)\\
& = \frac{\partial}{\partial \mb{u}}\,\begin{pmatrix}u_{1} \delta_{1} +
u_{2} \delta_{2} \\\mleft( - u_{2} \delta_{1} \mright) + u_{1}
\delta_{2} \end{pmatrix}\\
& = \begin{pmatrix}\delta_{1} & \delta_{2}\\\delta_{2} & -
\delta_{1}\end{pmatrix}\\
& = \bs{\Delta}\\
& = \overline{\bs{\Omega}}\mleft( \bs{\delta} \mright)
\end{align}
we get
\begin{align}
&\phantom{{}={}} \frac{\partial}{\partial \mb{u}}\,f\mleft( \mb{p},\mb{p}' \mright) \nonumber \\
& = \mleft( \frac{\partial}{\partial \mb{d}}\,f\mleft( \mb{p},\mb{p}'
\mright) \mright) \mleft( \frac{\partial}{\partial
\mb{u}}\,\mb{d}\mleft( \mb{x},\mb{u},\mb{x}' \mright) \mright)\\
& = \mleft( \mb{d}\mleft( \mb{x},\mb{u},\mb{x}' \mright) - \mb{r} \mright)^T
\mb{T}^{-1} \mleft( \frac{\partial}{\partial \mb{u}}\,\mb{d}\mleft(
\mb{x},\mb{u},\mb{x}' \mright) \mright)\\
& = \mleft( \mb{d}\mleft( \mb{x},\mb{u},\mb{x}' \mright) - \mb{r} \mright)^T
\mb{T}^{-1} \bs{\Delta}\\
& = \mleft( \mb{d}\mleft( \mb{x},\mb{u},\mb{x}' \mright) - \mb{r} \mright)^T
\mb{T}^{-1} \begin{pmatrix}\delta_{1} & \delta_{2}\\\delta_{2} & -
\delta_{1}\end{pmatrix}\\
& = \mleft( \mb{d}\mleft( \mb{x},\mb{u},\mb{x}' \mright) - \mb{r} \mright)^T
\mb{T}^{-1} \begin{pmatrix}x'_{1} - x_{1} & x'_{2} - x_{2}\\x'_{2} -
x_{2} & - \mleft( x'_{1} - x_{1} \mright)\end{pmatrix}.
\end{align}

\subsection{First Derivatives of Distance Error}\label{sec_first_dist}

%
We use
\begin{align}
\bs{\delta}_0
&=
{\,\frac{\bs{\delta}}{\mleft\| \bs{\delta} \mright\|}\,}
\end{align}
to compute the first derivatives of the distance error
\eqref{eq_dist_error}:
\begin{align}
&\phantom{{}={}} \frac{\partial}{\partial \mb{x}}\,e\mleft( \mb{p},\mb{p}' \mright) \nonumber \\
& = \mleft( \frac{\partial}{\partial \bs{\delta}}\,e\mleft( \mb{p},\mb{p}'
\mright) \mright) \mleft( \frac{\partial}{\partial \mb{x}}\,\bs{\delta}
\mright)\\
& = \mleft( \frac{\partial}{\partial \bs{\delta}}\,\mleft[ \frac{1}{2}\,
{\,\frac{1}{\sigma_e}\,} \mleft\{ \mleft\| \bs{\delta} \mright\| -
\varrho \mright\}^2 \mright] \mright) \mleft( - \mb{I}_{2} \mright)\\
& = - {\,\frac{1}{\sigma_e}\,} \mleft( \mleft\| \bs{\delta} \mright\| -
\varrho \mright) {\,\frac{\bs{\delta}^T}{\mleft\| \bs{\delta}
\mright\|}\,}
\end{align}
\parbox{\textwidth}{\rule{\textwidth}{0.4pt}}
\begin{align}
&\phantom{{}={}} \frac{\partial}{\partial \mb{x}'}\,e\mleft( \mb{p},\mb{p}' \mright) \nonumber \\
& = \mleft( \frac{\partial}{\partial \bs{\delta}}\,e\mleft( \mb{p},\mb{p}'
\mright) \mright) \mleft( \frac{\partial}{\partial \mb{x}'}\,\bs{\delta}
\mright)\\
& = \mleft( \frac{\partial}{\partial \bs{\delta}}\,\mleft[ \frac{1}{2}\,
{\,\frac{1}{\sigma_e}\,} \mleft\{ \mleft\| \bs{\delta} \mright\| -
\varrho \mright\}^2 \mright] \mright) \mb{I}_{2}\\
& = {\,\frac{1}{\sigma_e}\,} \mleft( \mleft\| \bs{\delta} \mright\| -
\varrho \mright) {\,\frac{\bs{\delta}^T}{\mleft\| \bs{\delta}
\mright\|}\,}.
\end{align}

\subsection{First Derivatives of Rotation Error}\label{sec_first_rot}

%
We get the first derivatives of the rotation error \eqref{eq_rot_error}
and \eqref{eq_rot_error_norm} from the generic solutions in section
\ref{sec_first_genrot}. For the first form we obtain
\begin{align}
&\phantom{{}={}} \frac{\partial}{\partial \mb{u}}\,g\mleft( \mb{p},\mb{p}' \mright) \nonumber \\
& = \gamma {\,\frac{1}{\sigma^{2}}\,} \mleft( \bar{t}_1 \mleft\| \mb{u}'
\mright\| \mb{u}_0^T - \mb{u}'^T \mb{Q} \mright)
\end{align}
\parbox{\textwidth}{\rule{\textwidth}{0.4pt}}
\begin{align}
&\phantom{{}={}} \frac{\partial}{\partial \mb{u}'}\,g\mleft( \mb{p},\mb{p}' \mright) \nonumber \\
& = \gamma {\,\frac{1}{\sigma^{2}}\,} \mleft( \bar{t}_1 \mleft\| \mb{u}
\mright\| {{} {\mb{u}'_0}}^T - \mb{u}^T \mb{Q}^T \mright).
\end{align}
For the second form we get
\begin{align}
&\phantom{{}={}} \frac{\partial}{\partial \mb{u}}\,\bar{g}\mleft( \mb{p},\mb{p}' \mright) \nonumber \\
& = - \gamma {\,\frac{1}{\sigma^{2}}\,} {{} {\mb{u}'_0}}^T \mb{Q}
{\,\frac{1}{\mleft\| \mb{u} \mright\|}\,} \mleft( \mb{I}_{2} - \mb{u}_0
\mb{u}_0^T \mright)
\end{align}
\parbox{\textwidth}{\rule{\textwidth}{0.4pt}}
\begin{align}
&\phantom{{}={}} \frac{\partial}{\partial \mb{u}'}\,\bar{g}\mleft( \mb{p},\mb{p}'
\mright) \nonumber \\
& = - \gamma {\,\frac{1}{\sigma^{2}}\,} \mb{u}_0^T \mb{Q}^T
{\,\frac{1}{\mleft\| \mb{u}' \mright\|}\,} \mleft( \mb{I}_{2} - {{}
{\mb{u}'_0}} {{} {\mb{u}'_0}}^T \mright).
\end{align}

\subsection{First Derivatives of Home-Vector Error}\label{sec_first_home}

%
We compute the first derivatives of the home-vector error
\eqref{eq_homevector_error} and \eqref{eq_homevector_error_norm}. We use
\begin{align}
\bs{\delta}_0
&=
{\,\frac{\bs{\delta}}{\mleft\| \bs{\delta} \mright\|}\,}\\
\bs{\delta}
&=
\mb{x}' - \mb{x}\\
\frac{\partial}{\partial \mb{x}}\,\bs{\delta} & = \frac{\partial}{\partial \mb{x}}\,\mleft( \mb{x}' - \mb{x} \mright) = - \mb{I}_{2}\\
\frac{\partial}{\partial \mb{x}'}\,\bs{\delta} & = \frac{\partial}{\partial \mb{x}'}\,\mleft( \mb{x}' - \mb{x} \mright) = \mb{I}_{2}
\end{align}
and obtain for the first form
\begin{align}
&\phantom{{}={}} \frac{\partial}{\partial \mb{x}}\,h\mleft( \mb{p},\mb{p}' \mright) \nonumber \\
& = \mleft( \frac{\partial}{\partial \bs{\delta}_0}\,h\mleft( \mb{p},\mb{p}'
\mright) \mright) \mleft( \frac{\partial}{\partial
\bs{\delta}}\,\bs{\delta}_0 \mright) \mleft( \frac{\partial}{\partial
\mb{x}}\,\bs{\delta} \mright)\\
& = \mleft( \frac{\partial}{\partial \bs{\delta}_0}\,\mleft[ \gamma
{\,\frac{1}{\sigma_h^{2}}\,} \mleft\{ t_1 + \bar{t}_1 \mleft\| \mb{u}
\mright\| - \mleft( \bs{A} \mb{u} \mright)^T \bs{\delta}_0 \mright\}
\mright] \mright) \mleft( \frac{\partial}{\partial
\bs{\delta}}\,\bs{\delta}_0 \mright) \mleft( \frac{\partial}{\partial
\mb{x}}\,\bs{\delta} \mright)\\
& = \gamma {\,\frac{1}{\sigma_h^{2}}\,} \mb{u}^T \bs{A}^T \mleft(
{\,\frac{\mb{I}_{2}}{\mleft\| \bs{\delta} \mright\|}\,} -
{\,\frac{\bs{\delta} \bs{\delta}^T }{\mleft\| \bs{\delta}
\mright\|^{3}}\,} \mright)\\
& = \gamma {\,\frac{1}{\sigma_h^{2}}\,} \mb{u}^T \bs{A}^T
{\,\frac{1}{\mleft\| \bs{\delta} \mright\|}\,} \mleft( \mb{I}_{2} -
\bs{\delta}_0 \bs{\delta}_0^T \mright)
\end{align}
\parbox{\textwidth}{\rule{\textwidth}{0.4pt}}
\begin{align}
&\phantom{{}={}} \frac{\partial}{\partial \mb{x}'}\,h\mleft( \mb{p},\mb{p}' \mright) \nonumber \\
& = \mleft( \frac{\partial}{\partial \bs{\delta}_0}\,h\mleft( \mb{p},\mb{p}'
\mright) \mright) \mleft( \frac{\partial}{\partial
\bs{\delta}}\,\bs{\delta}_0 \mright) \mleft( \frac{\partial}{\partial
\mb{x}'}\,\bs{\delta} \mright)\\
& = \mleft( \frac{\partial}{\partial \bs{\delta}_0}\,\mleft[ \gamma
{\,\frac{1}{\sigma_h^{2}}\,} \mleft\{ t_1 + \bar{t}_1 \mleft\| \mb{u}
\mright\| - \mleft( \bs{A} \mb{u} \mright)^T \bs{\delta}_0 \mright\}
\mright] \mright) \mleft( \frac{\partial}{\partial
\bs{\delta}}\,\bs{\delta}_0 \mright) \mleft( \frac{\partial}{\partial
\mb{x}'}\,\bs{\delta} \mright)\\
& = - \gamma {\,\frac{1}{\sigma_h^{2}}\,} \mb{u}^T \bs{A}^T \mleft(
{\,\frac{\mb{I}_{2}}{\mleft\| \bs{\delta} \mright\|}\,} -
{\,\frac{\bs{\delta} \bs{\delta}^T }{\mleft\| \bs{\delta}
\mright\|^{3}}\,} \mright)\\
& = - \gamma {\,\frac{1}{\sigma_h^{2}}\,} \mb{u}^T \bs{A}^T
{\,\frac{1}{\mleft\| \bs{\delta} \mright\|}\,} \mleft( \mb{I}_{2} -
\bs{\delta}_0 \bs{\delta}_0^T \mright)
\end{align}
\parbox{\textwidth}{\rule{\textwidth}{0.4pt}}
\begin{align}
&\phantom{{}={}} \frac{\partial}{\partial \mb{u}}\,h\mleft( \mb{p},\mb{p}' \mright) \nonumber \\
& = \frac{\partial}{\partial \mb{u}}\,\mleft( \gamma
{\,\frac{1}{\sigma_h^{2}}\,} \mleft[ t_1 + \bar{t}_1 \mleft\| \mb{u}
\mright\| - \mleft\{ \bs{A} \mb{u} \mright\}^T
{\,\frac{\bs{\delta}}{\mleft\| \bs{\delta} \mright\|}\,} \mright]
\mright)\\
& = \frac{\partial}{\partial \mb{u}}\,\mleft( \gamma
{\,\frac{1}{\sigma_h^{2}}\,} \mleft[ t_1 + \bar{t}_1 \mleft\| \mb{u}
\mright\| - {\,\frac{\bs{\delta}^T}{\mleft\| \bs{\delta} \mright\|}\,}
\bs{A} \mb{u} \mright] \mright)\\
& = \gamma {\,\frac{1}{\sigma_h^{2}}\,} \mleft( \bar{t}_1
{\,\frac{\mb{u}^T}{\mleft\| \mb{u} \mright\|}\,} -
{\,\frac{\bs{\delta}^T}{\mleft\| \bs{\delta} \mright\|}\,} \bs{A}
\mright)\\
& = \gamma {\,\frac{1}{\sigma_h^{2}}\,} \mleft( \bar{t}_1 \mb{u}_0^T -
\bs{\delta}_0^T \bs{A} \mright).
\end{align}
For the second form we get
\begin{align}
&\phantom{{}={}} \frac{\partial}{\partial \mb{x}}\,\bar{h}\mleft( \mb{p},\mb{p}' \mright) \nonumber \\
& = \mleft( \frac{\partial}{\partial \bs{\delta}_0}\,\bar{h}\mleft(
\mb{p},\mb{p}' \mright) \mright) \mleft( \frac{\partial}{\partial
\bs{\delta}}\,\bs{\delta}_0 \mright) \mleft( \frac{\partial}{\partial
\mb{x}}\,\bs{\delta} \mright)\\
& = \mleft( \frac{\partial}{\partial \bs{\delta}_0}\,\mleft[ \gamma
{\,\frac{1}{\sigma_h^{2}}\,} \mleft\{ 1 - \mleft( \bs{A}
{\,\frac{\mb{u}}{\mleft\| \mb{u} \mright\|}\,} \mright)^T \bs{\delta}_0
\mright\} \mright] \mright) \mleft( \frac{\partial}{\partial
\bs{\delta}}\,\bs{\delta}_0 \mright) \mleft( \frac{\partial}{\partial
\mb{x}}\,\bs{\delta} \mright)\\
& = \gamma {\,\frac{1}{\sigma_h^{2}}\,} {\,\frac{\mb{u}^T}{\mleft\| \mb{u}
\mright\|}\,} \bs{A}^T \mleft( {\,\frac{\mb{I}_{2}}{\mleft\| \bs{\delta}
\mright\|}\,} - {\,\frac{\bs{\delta} \bs{\delta}^T }{\mleft\|
\bs{\delta} \mright\|^{3}}\,} \mright)\\
& = \gamma {\,\frac{1}{\sigma_h^{2}}\,} \mb{u}_0^T \bs{A}^T
{\,\frac{1}{\mleft\| \bs{\delta} \mright\|}\,} \mleft( \mb{I}_{2} -
\bs{\delta}_0 \bs{\delta}_0^T \mright)
\end{align}
\parbox{\textwidth}{\rule{\textwidth}{0.4pt}}
\begin{align}
&\phantom{{}={}} \frac{\partial}{\partial \mb{x}'}\,\bar{h}\mleft( \mb{p},\mb{p}'
\mright) \nonumber \\
& = \mleft( \frac{\partial}{\partial \bs{\delta}_0}\,\bar{h}\mleft(
\mb{p},\mb{p}' \mright) \mright) \mleft( \frac{\partial}{\partial
\bs{\delta}}\,\bs{\delta}_0 \mright) \mleft( \frac{\partial}{\partial
\mb{x}'}\,\bs{\delta} \mright)\\
& = \mleft( \frac{\partial}{\partial \bs{\delta}_0}\,\mleft[ \gamma
{\,\frac{1}{\sigma_h^{2}}\,} \mleft\{ 1 - \mleft( \bs{A}
{\,\frac{\mb{u}}{\mleft\| \mb{u} \mright\|}\,} \mright)^T \bs{\delta}_0
\mright\} \mright] \mright) \mleft( \frac{\partial}{\partial
\bs{\delta}}\,\bs{\delta}_0 \mright) \mleft( \frac{\partial}{\partial
\mb{x}'}\,\bs{\delta} \mright)\\
& = - \gamma {\,\frac{1}{\sigma_h^{2}}\,} {\,\frac{\mb{u}^T}{\mleft\| \mb{u}
\mright\|}\,} \bs{A}^T \mleft( {\,\frac{\mb{I}_{2}}{\mleft\| \bs{\delta}
\mright\|}\,} - {\,\frac{\bs{\delta} \bs{\delta}^T }{\mleft\|
\bs{\delta} \mright\|^{3}}\,} \mright)\\
& = - \gamma {\,\frac{1}{\sigma_h^{2}}\,} \mb{u}_0^T \bs{A}^T
{\,\frac{1}{\mleft\| \bs{\delta} \mright\|}\,} \mleft( \mb{I}_{2} -
\bs{\delta}_0 \bs{\delta}_0^T \mright)
\end{align}
\parbox{\textwidth}{\rule{\textwidth}{0.4pt}}
\begin{align}
&\phantom{{}={}} \frac{\partial}{\partial \mb{u}}\,\bar{h}\mleft( \mb{p},\mb{p}' \mright) \nonumber \\
& = \frac{\partial}{\partial \mb{u}}\,\mleft( \gamma
{\,\frac{1}{\sigma_h^{2}}\,} \mleft[ 1 - \mleft\{ \bs{A}
{\,\frac{\mb{u}}{\mleft\| \mb{u} \mright\|}\,} \mright\}^T
{\,\frac{\bs{\delta}}{\mleft\| \bs{\delta} \mright\|}\,} \mright]
\mright)\\
& = \frac{\partial}{\partial \mb{u}}\,\mleft( \gamma
{\,\frac{1}{\sigma_h^{2}}\,} \mleft[ 1 -
{\,\frac{\bs{\delta}^T}{\mleft\| \bs{\delta} \mright\|}\,} \bs{A}
{\,\frac{\mb{u}}{\mleft\| \mb{u} \mright\|}\,} \mright] \mright)\\
& = - \gamma {\,\frac{1}{\sigma_h^{2}}\,} {\,\frac{\bs{\delta}^T}{\mleft\|
\bs{\delta} \mright\|}\,} \bs{A} \mleft( {\,\frac{\mb{I}_{2}}{\mleft\|
\mb{u} \mright\|}\,} - {\,\frac{\mb{u} \mb{u}^T }{\mleft\| \mb{u}
\mright\|^{3}}\,} \mright)\\
& = - \gamma {\,\frac{1}{\sigma_h^{2}}\,} \bs{\delta}_0^T \bs{A}
{\,\frac{1}{\mleft\| \mb{u} \mright\|}\,} \mleft( \mb{I}_{2} - \mb{u}_0
\mb{u}_0^T \mright).
\end{align}

\subsection{First Derivatives of Compass Error}\label{sec_first_compass}

%
We get the first derivatives of the compass error
\eqref{eq_compass_error} and \eqref{eq_compass_error_norm} from the
generic solutions in section \ref{sec_first_genrot}. For the first form
we obtain
\begin{align}
&\phantom{{}={}} \frac{\partial}{\partial \mb{u}}\,c\mleft( \mb{p},\mb{p}' \mright) \nonumber \\
& = \gamma {\,\frac{1}{\sigma_c^{2}}\,} \mleft( \bar{t}_1 \mleft\| \mb{u}'
\mright\| \mb{u}_0^T - \mb{u}'^T \bs{\Psi} \mright)
\end{align}
\parbox{\textwidth}{\rule{\textwidth}{0.4pt}}
\begin{align}
&\phantom{{}={}} \frac{\partial}{\partial \mb{u}'}\,c\mleft( \mb{p},\mb{p}' \mright) \nonumber \\
& = \gamma {\,\frac{1}{\sigma_c^{2}}\,} \mleft( \bar{t}_1 \mleft\| \mb{u}
\mright\| {{} {\mb{u}'_0}}^T - \mb{u}^T \bs{\Psi}^T \mright).
\end{align}
For the second form we get
\begin{align}
&\phantom{{}={}} \frac{\partial}{\partial \mb{u}}\,\bar{c}\mleft( \mb{p},\mb{p}' \mright) \nonumber \\
& = - \gamma {\,\frac{1}{\sigma_c^{2}}\,} {{} {\mb{u}'_0}}^T \bs{\Psi}
{\,\frac{1}{\mleft\| \mb{u} \mright\|}\,} \mleft( \mb{I}_{2} - \mb{u}_0
\mb{u}_0^T \mright)
\end{align}
\parbox{\textwidth}{\rule{\textwidth}{0.4pt}}
\begin{align}
&\phantom{{}={}} \frac{\partial}{\partial \mb{u}'}\,\bar{c}\mleft( \mb{p},\mb{p}'
\mright) \nonumber \\
& = - \gamma {\,\frac{1}{\sigma_c^{2}}\,} \mb{u}_0^T \bs{\Psi}^T
{\,\frac{1}{\mleft\| \mb{u}' \mright\|}\,} \mleft( \mb{I}_{2} - {{}
{\mb{u}'_0}} {{} {\mb{u}'_0}}^T \mright).
\end{align}

\subsection{Second Derivatives}\label{sec_second}

%
In the following, we exploit the fact that second derivatives w.r.t. the
same (scalar) variables but in different order are identical, so we can
reduce the number of derivatives to compute. When the final matrices are
formed, the sub-blocks of the missing derivatives are the transposed
sub-blocks of the ones which are specified.

Zero second derivatives are not listed.

Note that (second) derivatives of first derivatives with respect to the
same vector variable are symmetric (which can be used to check the
results).

For each cost function, we first determine the transposed of the first
derivative (since the first derivatives are row vectors, but the partial
derivative expects a column vector).

In intermediate steps, the orientation matrix $\mb{U}$ is expressed by
$\mb{u}$ through $\mb{U} = \bs{\Omega}\mleft( \mb{u} \mright)$ (see
section \ref{sec_orvec}). We also use
\begin{align}
\bs{\Delta} & = \overline{\bs{\Omega}}\mleft( \bs{\delta} \mright) = \begin{pmatrix}\delta_{1} & \delta_{2}\\\delta_{2} & -
\delta_{1}\end{pmatrix}.
\end{align}

\subsection{Second Derivatives of Generic Rotational Cost Function}\label{sec_second_genrot}

%
For the first form we get
\begin{align}
&\phantom{{}={}} \frac{\partial}{\partial \mb{u}}\,\mleft( \frac{\partial}{\partial
\mb{u}}\,s\mleft( \bs{\Phi},\mb{p},\mb{p}' \mright) \mright)^T \nonumber \\
& = \frac{\partial}{\partial \mb{u}}\,\mleft( \bar{t}_1 \mleft\| \mb{u}'
\mright\| {\,\frac{\mb{u}}{\mleft\| \mb{u} \mright\|}\,} - \bs{\Phi}^T
\mb{u}' \mright)\\
& = \bar{t}_1 {\,\frac{\mleft\| \mb{u}' \mright\|}{\mleft\| \mb{u}
\mright\|}\,} \mleft( \mb{I}_{2} - \mb{u}_0 \mb{u}_0^T \mright)
\end{align}
\parbox{\textwidth}{\rule{\textwidth}{0.4pt}}
\begin{align}
&\phantom{{}={}} \frac{\partial}{\partial \mb{u}'}\,\mleft( \frac{\partial}{\partial
\mb{u}}\,s\mleft( \bs{\Phi},\mb{p},\mb{p}' \mright) \mright)^T \nonumber \\
& = \frac{\partial}{\partial \mb{u}'}\,\mleft( \bar{t}_1 \mleft\| \mb{u}'
\mright\| {\,\frac{\mb{u}}{\mleft\| \mb{u} \mright\|}\,} - \bs{\Phi}^T
\mb{u}' \mright)\\
& = \bar{t}_1 {\,\frac{\mb{u}}{\mleft\| \mb{u} \mright\|}\,}
{\,\frac{\mb{u}'^T}{\mleft\| \mb{u}' \mright\|}\,} - \bs{\Phi}^T\\
& = \bar{t}_1 \mb{u}_0 {{} {\mb{u}'_0}}^T - \bs{\Phi}^T
\end{align}
\parbox{\textwidth}{\rule{\textwidth}{0.4pt}}
\begin{align}
&\phantom{{}={}} \frac{\partial}{\partial \mb{u}'}\,\mleft( \frac{\partial}{\partial
\mb{u}'}\,s\mleft( \bs{\Phi},\mb{p},\mb{p}' \mright) \mright)^T \nonumber \\
& = \frac{\partial}{\partial \mb{u}'}\,\mleft( \bar{t}_1 \mleft\| \mb{u}
\mright\| {\,\frac{\mb{u}'}{\mleft\| \mb{u}' \mright\|}\,} - \bs{\Phi}
\mb{u} \mright)\\
& = \bar{t}_1 {\,\frac{\mleft\| \mb{u} \mright\|}{\mleft\| \mb{u}'
\mright\|}\,} \mleft( \mb{I}_{2} - {{} {\mb{u}'_0}} {{} {\mb{u}'_0}}^T
\mright).
\end{align}
For the second form we get
\begin{align}
&\phantom{{}={}} \frac{\partial}{\partial \mb{u}}\,\mleft( \frac{\partial}{\partial
\mb{u}}\,\bar{s}\mleft( \bs{\Phi},\mb{p},\mb{p}' \mright) \mright)^T \nonumber \\
& = - \frac{\partial}{\partial \mb{u}}\,\mleft( \mleft[
{\,\frac{\mb{I}_{2}}{\mleft\| \mb{u} \mright\|}\,} - {\,\frac{\mb{u}
\mb{u}^T }{\mleft\| \mb{u} \mright\|^{3}}\,} \mright] \bs{\Phi}^T {{}
{\mb{u}'_0}} \mright)\\
& = {\,\frac{1}{\mleft\| \mb{u} \mright\|^{2}}\,} \mleft( \bs{\Phi}^T {{}
{\mb{u}'_0}} \mb{u}_0^T + \mleft[ \mb{u}_0^T \bs{\Phi}^T {{}
{\mb{u}'_0}} \mright] \mleft[ \mb{I}_{2} - 3 \mb{u}_0 \mb{u}_0^T
\mright] + \mb{u}_0 {{} {\mb{u}'_0}}^T \bs{\Phi} \mright)
\end{align}
\parbox{\textwidth}{\rule{\textwidth}{0.4pt}}
\begin{align}
&\phantom{{}={}} \frac{\partial}{\partial \mb{u}'}\,\mleft( \frac{\partial}{\partial
\mb{u}}\,\bar{s}\mleft( \bs{\Phi},\mb{p},\mb{p}' \mright) \mright)^T \nonumber \\
& = - \frac{\partial}{\partial \mb{u}'}\,\mleft( \mleft[
{\,\frac{\mb{I}_{2}}{\mleft\| \mb{u} \mright\|}\,} - {\,\frac{\mb{u}
\mb{u}^T }{\mleft\| \mb{u} \mright\|^{3}}\,} \mright] \bs{\Phi}^T {{}
{\mb{u}'_0}} \mright)\\
& = - \mleft( {\,\frac{\mb{I}_{2}}{\mleft\| \mb{u} \mright\|}\,} -
{\,\frac{\mb{u} \mb{u}^T }{\mleft\| \mb{u} \mright\|^{3}}\,} \mright)
\bs{\Phi}^T \mleft( {\,\frac{\mb{I}_{2}}{\mleft\| \mb{u}' \mright\|}\,}
- {\,\frac{\mb{u}' \mb{u}'^T }{\mleft\| \mb{u}' \mright\|^{3}}\,}
\mright)\\
& = - {\,\frac{1}{\mleft\| \mb{u} \mright\| \mleft\| \mb{u}' \mright\| }\,}
\mleft( \mb{I}_{2} - \mb{u}_0 \mb{u}_0^T \mright) \bs{\Phi}^T \mleft(
\mb{I}_{2} - {{} {\mb{u}'_0}} {{} {\mb{u}'_0}}^T \mright)
\end{align}
\parbox{\textwidth}{\rule{\textwidth}{0.4pt}}
\begin{align}
&\phantom{{}={}} \frac{\partial}{\partial \mb{u}'}\,\mleft( \frac{\partial}{\partial
\mb{u}'}\,\bar{s}\mleft( \bs{\Phi},\mb{p},\mb{p}' \mright) \mright)^T \nonumber \\
& = - \frac{\partial}{\partial \mb{u}'}\,\mleft( \mleft[
{\,\frac{\mb{I}_{2}}{\mleft\| \mb{u}' \mright\|}\,} - {\,\frac{\mb{u}'
\mb{u}'^T }{\mleft\| \mb{u}' \mright\|^{3}}\,} \mright] \bs{\Phi}
\mb{u}_0 \mright)\\
& = {\,\frac{1}{\mleft\| \mb{u}' \mright\|^{2}}\,} \mleft( \bs{\Phi}
\mb{u}_0 {{} {\mb{u}'_0}}^T + \mleft[ {{} {\mb{u}'_0}}^T \bs{\Phi}
\mb{u}_0 \mright] \mleft[ \mb{I}_{2} - 3 {{} {\mb{u}'_0}} {{}
{\mb{u}'_0}}^T \mright] + {{} {\mb{u}'_0}} \mb{u}_0^T \bs{\Phi}^T
\mright).
\end{align}

\subsection{Second Derivatives of Translation Error}\label{sec_second_trans}

%
\textbf{For first derivative with respect to $\mb{x}$:}
\begin{align}
&\phantom{{}={}} \mleft( \frac{\partial}{\partial \mb{x}}\,f\mleft( \mb{p},\mb{p}^T
\mright) \mright)^T \nonumber \\
& = \mleft( - \mleft[ \mb{d}\mleft( \mb{x},\mb{u},\mb{x}' \mright) - \mb{r}
\mright]^T \mb{T}^{-1} \mb{U}^T \mright)^T\\
& = - \mb{U} \mb{T}^{-1} \mleft( \mb{d}\mleft( \mb{x},\mb{u},\mb{x}'
\mright) - \mb{r} \mright)\\
& = - \mb{U} \mb{T}^{-1} \mleft( \mb{U}^T \mleft[ \mb{x}' - \mb{x} \mright]
- \mb{r} \mright).
\end{align}
\textbf{Second derivatives:}
\begin{align}
&\phantom{{}={}} \frac{\partial}{\partial \mb{x}}\,\mleft( \frac{\partial}{\partial
\mb{x}}\,f\mleft( \mb{p},\mb{p}^T \mright) \mright)^T \nonumber \\
& = \frac{\partial}{\partial \mb{x}}\,\mleft( - \mb{U} \mb{T}^{-1} \mleft[
\mb{U}^T \mleft\{ \mb{x}' - \mb{x} \mright\} - \mb{r} \mright] \mright)\\
& = \mb{U} \mb{T}^{-1} \mb{U}^T
\end{align}
\parbox{\textwidth}{\rule{\textwidth}{0.4pt}}
\begin{align}
&\phantom{{}={}} \frac{\partial}{\partial \mb{x}'}\,\mleft( \frac{\partial}{\partial
\mb{x}}\,f\mleft( \mb{p},\mb{p}^T \mright) \mright)^T \nonumber \\
& = \frac{\partial}{\partial \mb{x}'}\,\mleft( - \mb{U} \mb{T}^{-1} \mleft[
\mb{U}^T \mleft\{ \mb{x}' - \mb{x} \mright\} - \mb{r} \mright] \mright)\\
& = - \mb{U} \mb{T}^{-1} \mb{U}^T
\end{align}
\parbox{\textwidth}{\rule{\textwidth}{0.4pt}}
\begin{align}
&\phantom{{}={}} \frac{\partial}{\partial \mb{u}}\,\mleft( \frac{\partial}{\partial
\mb{x}}\,f\mleft( \mb{p},\mb{p}^T \mright) \mright)^T \nonumber \\
& = \frac{\partial}{\partial \mb{u}}\,\mleft( - \mb{U} \mb{T}^{-1} \mleft[
\mb{U}^T \bs{\delta} - \mb{r} \mright] \mright)\\
& = - \frac{\partial}{\partial \mb{u}}\,\mleft( \mb{U} \mb{T}^{-1} \mb{U}^T
\bs{\delta} - \mb{U} \mb{T}^{-1} \mb{r} \mright)\\
& = - \frac{\partial}{\partial \mb{u}}\,\mleft( \begin{pmatrix}\mb{u}^T
\mb{M} \\\mb{u}^T \mb{N} \end{pmatrix} \mb{T}^{-1} \begin{pmatrix}\mb{M}
\mb{u} & \mb{N} \mb{u} \end{pmatrix} \bs{\delta} -
\begin{pmatrix}\mb{u}^T \mb{M} \\\mb{u}^T \mb{N} \end{pmatrix}
\mb{T}^{-1} \mb{r} \mright)\\
& = - \frac{\partial}{\partial \mb{u}}\,\mleft( \begin{pmatrix}\mb{u}^T
\mb{M} \mb{T}^{-1} \\\mb{u}^T \mb{N} \mb{T}^{-1} \end{pmatrix}
\begin{pmatrix}\mb{M} \mb{u} & \mb{N} \mb{u} \end{pmatrix} \bs{\delta} -
\begin{pmatrix}\mb{u}^T \mb{M} \mb{T}^{-1} \mb{r} \\\mb{u}^T \mb{N}
\mb{T}^{-1} \mb{r} \end{pmatrix} \mright)\\
& = - \frac{\partial}{\partial \mb{u}}\,\mleft( \begin{pmatrix}\mb{u}^T
\mb{M} \mb{T}^{-1} \mb{M} \mb{u} & \mb{u}^T \mb{M} \mb{T}^{-1} \mb{N}
\mb{u} \\\mb{u}^T \mb{N} \mb{T}^{-1} \mb{M} \mb{u} & \mb{u}^T \mb{N}
\mb{T}^{-1} \mb{N} \mb{u} \end{pmatrix}
\begin{pmatrix}\delta_{1}\\\delta_{2}\end{pmatrix} -
\begin{pmatrix}\mb{u}^T \mb{M} \mb{T}^{-1} \mb{r} \\\mb{u}^T \mb{N}
\mb{T}^{-1} \mb{r} \end{pmatrix} \mright)\\
& = - \frac{\partial}{\partial \mb{u}}\,\mleft( \begin{pmatrix}\mb{u}^T
\mb{M} \mb{T}^{-1} \mb{M} \mb{u} \delta_{1} + \mb{u}^T \mb{M}
\mb{T}^{-1} \mb{N} \mb{u} \delta_{2} \\\mb{u}^T \mb{N} \mb{T}^{-1}
\mb{M} \mb{u} \delta_{1} + \mb{u}^T \mb{N} \mb{T}^{-1} \mb{N} \mb{u}
\delta_{2} \end{pmatrix} - \begin{pmatrix}\mb{u}^T \mb{M} \mb{T}^{-1}
\mb{r} \\\mb{u}^T \mb{N} \mb{T}^{-1} \mb{r} \end{pmatrix} \mright)\\
& = - \mleft( \begin{pmatrix}\frac{\partial}{\partial \mb{u}}\,\mleft(
\mb{u}^T \mb{M} \mb{T}^{-1} \mb{M} \mb{u} \delta_{1} + \mb{u}^T \mb{M}
\mb{T}^{-1} \mb{N} \mb{u} \delta_{2} \mright)\\\frac{\partial}{\partial
\mb{u}}\,\mleft( \mb{u}^T \mb{N} \mb{T}^{-1} \mb{M} \mb{u} \delta_{1} +
\mb{u}^T \mb{N} \mb{T}^{-1} \mb{N} \mb{u} \delta_{2}
\mright)\end{pmatrix} - \begin{pmatrix}\frac{\partial}{\partial
\mb{u}}\,\mleft( \mb{u}^T \mb{M} \mb{T}^{-1} \mb{r}
\mright)\\\frac{\partial}{\partial \mb{u}}\,\mleft( \mb{u}^T \mb{N}
\mb{T}^{-1} \mb{r} \mright)\end{pmatrix} \mright)\\
& = - \mleft( \begin{pmatrix}2 \mb{u}^T \mb{M} \mb{T}^{-1} \mb{M} \delta_{1}
+ \mb{u}^T \mb{M} \mb{T}^{-1} \mb{N} \delta_{2} + \mb{u}^T \mb{N}
\mb{T}^{-1} \mb{M} \delta_{2} \\\mb{u}^T \mb{N} \mb{T}^{-1} \mb{M}
\delta_{1} + \mb{u}^T \mb{M} \mb{T}^{-1} \mb{N} \delta_{1} + 2 \mb{u}^T
\mb{N} \mb{T}^{-1} \mb{N} \delta_{2} \end{pmatrix} -
\begin{pmatrix}\mb{r}^T \mb{T}^{-1} \mb{M} \\\mb{r}^T \mb{T}^{-1} \mb{N}
\end{pmatrix} \mright)\\
& = - \mleft( \begin{pmatrix}\mb{u}^T \mb{M} \mb{T}^{-1} \mleft( \mb{M}
\delta_{1} + \mb{N} \delta_{2} \mright) + \mb{u}^T \mleft( \mb{M}
\delta_{1} + \mb{N} \delta_{2} \mright) \mb{T}^{-1} \mb{M} \\\mb{u}^T
\mb{N} \mb{T}^{-1} \mleft( \mb{M} \delta_{1} + \mb{N} \delta_{2}
\mright) + \mb{u}^T \mleft( \mb{M} \delta_{1} + \mb{N} \delta_{2}
\mright) \mb{T}^{-1} \mb{N} \end{pmatrix} - \begin{pmatrix}\mb{r}^T
\mb{T}^{-1} \mb{M} \\\mb{r}^T \mb{T}^{-1} \mb{N} \end{pmatrix} \mright)\\
& = - \mleft( \begin{pmatrix}\mb{u}^T \mb{M} \mb{T}^{-1} \bs{\Delta}
\\\mb{u}^T \mb{N} \mb{T}^{-1} \bs{\Delta} \end{pmatrix} +
\begin{pmatrix}\mb{u}^T \bs{\Delta} \mb{T}^{-1} \mb{M} \\\mb{u}^T
\bs{\Delta} \mb{T}^{-1} \mb{N} \end{pmatrix} - \begin{pmatrix}\mb{r}^T
\mb{T}^{-1} \mb{M} \\\mb{r}^T \mb{T}^{-1} \mb{N} \end{pmatrix} \mright)\\
& = - \mleft( \begin{pmatrix}\mb{u}^T \mb{M} \\\mb{u}^T \mb{N} \end{pmatrix}
\mb{T}^{-1} \bs{\Delta} + \bs{\Omega}\mleft( \mb{T}^{-1} \bs{\Delta}
\mb{u} \mright) - \begin{pmatrix}\mb{r}^T \mb{T}^{-1} \mb{M} \\\mb{r}^T
\mb{T}^{-1} \mb{N} \end{pmatrix} \mright)\\
& = - \mleft( \mb{U} \mb{T}^{-1} \bs{\Delta} + \bs{\Omega}\mleft(
\mb{T}^{-1} \bs{\Delta} \mb{u} \mright) - \bs{\Omega}\mleft( \mb{T}^{-1}
\mb{r} \mright) \mright)\\
& = - \mleft( \mb{U} \mb{T}^{-1} \bs{\Delta} + \bs{\Omega}\mleft(
\mb{T}^{-1} \bs{\Delta} \mb{u} - \mb{T}^{-1} \mb{r} \mright) \mright)\\
& = - \mleft( \mb{U} \mb{T}^{-1} \bs{\Delta} + \bs{\Omega}\mleft(
\mb{T}^{-1} \mleft[ \bs{\Delta} \mb{u} - \mb{r} \mright] \mright)
\mright).
\end{align}
\textbf{For first derivative with respect to $\mb{x}'$:}
\begin{align}
&\phantom{{}={}} \mleft( \frac{\partial}{\partial \mb{x}'}\,f\mleft( \mb{p},\mb{p}^T
\mright) \mright)^T \nonumber \\
& = \mb{U} \mb{T}^{-1} \mleft( \mb{U}^T \mleft[ \mb{x}' - \mb{x} \mright] -
\mb{r} \mright).
\end{align}
\textbf{Second derivatives:}
\begin{align}
&\phantom{{}={}} \frac{\partial}{\partial \mb{x}'}\,\mleft( \frac{\partial}{\partial
\mb{x}'}\,f\mleft( \mb{p},\mb{p}^T \mright) \mright)^T \nonumber \\
& = \frac{\partial}{\partial \mb{x}'}\,\mleft( \mb{U} \mb{T}^{-1} \mleft[
\mb{U}^T \mleft\{ \mb{x}' - \mb{x} \mright\} - \mb{r} \mright] \mright)\\
& = \mb{U} \mb{T}^{-1} \mb{U}^T
\end{align}
\parbox{\textwidth}{\rule{\textwidth}{0.4pt}}
\begin{align}
&\phantom{{}={}} \frac{\partial}{\partial \mb{u}}\,\mleft( \frac{\partial}{\partial
\mb{x}'}\,f\mleft( \mb{p},\mb{p}^T \mright) \mright)^T \nonumber \\
& = \frac{\partial}{\partial \mb{u}}\,\mleft( \mb{U} \mb{T}^{-1} \mleft[
\mb{U}^T \bs{\delta} - \mb{r} \mright] \mright)\\
& = \mb{U} \mb{T}^{-1} \bs{\Delta} + \bs{\Omega}\mleft( \mb{T}^{-1} \mleft(
\bs{\Delta} \mb{u} - \mb{r} \mright) \mright).
\end{align}
\textbf{For first derivative with respect to $\mb{u}$:}

Derived in a form required for derivative with respect to $\mb{u}$:
\begin{align}
&\phantom{{}={}} \mleft( \frac{\partial}{\partial \mb{u}}\,f\mleft( \mb{p},\mb{p}'
\mright) \mright)^T \nonumber \\
& = \begin{pmatrix}\delta_{1} & \delta_{2}\\\delta_{2} & -
\delta_{1}\end{pmatrix} \mb{T}^{-1} \mleft( \mb{U}^T \bs{\delta} -
\mb{r} \mright)\\
& = \bs{\Delta} \mb{T}^{-1} \mb{U}^T \bs{\delta} - \bs{\Delta} \mb{T}^{-1}
\mb{r}\\
& = \bs{\Delta} \mb{T}^{-1} \begin{pmatrix}\mb{M} \mb{u} & \mb{N} \mb{u}
\end{pmatrix} \begin{pmatrix}\delta_{1}\\\delta_{2}\end{pmatrix} -
\bs{\Delta} \mb{T}^{-1} \mb{r}\\
& = \mleft( \bs{\Delta} \mb{T}^{-1} \mb{M} \mb{u} \delta_{1} + \bs{\Delta}
\mb{T}^{-1} \mb{N} \mb{u} \delta_{2} \mright) - \bs{\Delta} \mb{T}^{-1}
\mb{r}.
\end{align}
\textbf{Second derivative:}
\begin{align}
&\phantom{{}={}} \frac{\partial}{\partial \mb{u}}\,\mleft( \frac{\partial}{\partial
\mb{u}}\,f\mleft( \mb{p},\mb{p}' \mright) \mright)^T \nonumber \\
& = \frac{\partial}{\partial \mb{u}}\,\mleft( \bs{\Delta} \mb{T}^{-1} \mb{M}
\mb{u} \delta_{1} + \bs{\Delta} \mb{T}^{-1} \mb{N} \mb{u} \delta_{2} -
\bs{\Delta} \mb{T}^{-1} \mb{r} \mright)\\
& = \bs{\Delta} \mb{T}^{-1} \mb{M} \delta_{1} + \bs{\Delta} \mb{T}^{-1}
\mb{N} \delta_{2}\\
& = \bs{\Delta} \mb{T}^{-1} \mleft( \mb{M} \delta_{1} + \mb{N} \delta_{2}
\mright)\\
& = \bs{\Delta} \mb{T}^{-1} \mleft( \begin{pmatrix}1 & 0\\0 & -
1\end{pmatrix} \delta_{1} + \begin{pmatrix}0 & 1\\1 & 0\end{pmatrix}
\delta_{2} \mright)\\
& = \bs{\Delta} \mb{T}^{-1} \bs{\Delta}.
\end{align}

\subsection{Second Derivatives of Distance Error}\label{sec_second_dist}

%
\begin{align}
&\phantom{{}={}} \frac{\partial}{\partial \mb{x}}\,\mleft( \frac{\partial}{\partial
\mb{x}}\,e\mleft( \mb{p},\mb{p}' \mright) \mright)^T \nonumber \\
& = \frac{\partial}{\partial \mb{x}}\,\mleft( - {\,\frac{1}{\sigma_e}\,}
\mleft[ \mleft\| \bs{\delta} \mright\| - \varrho \mright]
{\,\frac{\bs{\delta}}{\mleft\| \bs{\delta} \mright\|}\,} \mright)\\
& = \frac{\partial}{\partial \mb{x}}\,\mleft( - {\,\frac{1}{\sigma_e}\,}
\mleft[ \bs{\delta} - \varrho {\,\frac{\bs{\delta}}{\mleft\| \bs{\delta}
\mright\|}\,} \mright] \mright)\\
& = {\,\frac{1}{\sigma_e}\,} \mleft( \mb{I}_{2} - \varrho \mleft[
{\,\frac{\mb{I}_{2}}{\mleft\| \bs{\delta} \mright\|}\,} -
{\,\frac{\bs{\delta} \bs{\delta}^T }{\mleft\| \bs{\delta}
\mright\|^{3}}\,} \mright] \mright)\\
& = {\,\frac{1}{\sigma_e}\,} \mleft( \mb{I}_{2} - \varrho
{\,\frac{1}{\mleft\| \bs{\delta} \mright\|}\,} \mleft[ \mb{I}_{2} -
\bs{\delta}_0 \bs{\delta}_0^T \mright] \mright)
\end{align}
\parbox{\textwidth}{\rule{\textwidth}{0.4pt}}
\begin{align}
&\phantom{{}={}} \frac{\partial}{\partial \mb{x}'}\,\mleft( \frac{\partial}{\partial
\mb{x}}\,e\mleft( \mb{p},\mb{p}' \mright) \mright)^T \nonumber \\
& = \frac{\partial}{\partial \mb{x}'}\,\mleft( - {\,\frac{1}{\sigma_e}\,}
\mleft[ \mleft\| \bs{\delta} \mright\| - \varrho \mright]
{\,\frac{\bs{\delta}}{\mleft\| \bs{\delta} \mright\|}\,} \mright)\\
& = \frac{\partial}{\partial \mb{x}'}\,\mleft( - {\,\frac{1}{\sigma_e}\,}
\mleft[ \bs{\delta} - \varrho {\,\frac{\bs{\delta}}{\mleft\| \bs{\delta}
\mright\|}\,} \mright] \mright)\\
& = - {\,\frac{1}{\sigma_e}\,} \mleft( \mb{I}_{2} - \varrho \mleft[
{\,\frac{\mb{I}_{2}}{\mleft\| \bs{\delta} \mright\|}\,} -
{\,\frac{\bs{\delta} \bs{\delta}^T }{\mleft\| \bs{\delta}
\mright\|^{3}}\,} \mright] \mright)\\
& = - {\,\frac{1}{\sigma_e}\,} \mleft( \mb{I}_{2} - \varrho
{\,\frac{1}{\mleft\| \bs{\delta} \mright\|}\,} \mleft[ \mb{I}_{2} -
\bs{\delta}_0 \bs{\delta}_0^T \mright] \mright)
\end{align}
\parbox{\textwidth}{\rule{\textwidth}{0.4pt}}
\begin{align}
&\phantom{{}={}} \frac{\partial}{\partial \mb{x}'}\,\mleft( \frac{\partial}{\partial
\mb{x}'}\,e\mleft( \mb{p},\mb{p}' \mright) \mright)^T \nonumber \\
& = \frac{\partial}{\partial \mb{x}'}\,\mleft( {\,\frac{1}{\sigma_e}\,}
\mleft[ \mleft\| \bs{\delta} \mright\| - \varrho \mright]
{\,\frac{\bs{\delta}}{\mleft\| \bs{\delta} \mright\|}\,} \mright)\\
& = \frac{\partial}{\partial \mb{x}'}\,\mleft( {\,\frac{1}{\sigma_e}\,}
\mleft[ \bs{\delta} - \varrho {\,\frac{\bs{\delta}}{\mleft\| \bs{\delta}
\mright\|}\,} \mright] \mright)\\
& = {\,\frac{1}{\sigma_e}\,} \mleft( \mb{I}_{2} - \varrho \mleft[
{\,\frac{\mb{I}_{2}}{\mleft\| \bs{\delta} \mright\|}\,} -
{\,\frac{\bs{\delta} \bs{\delta}^T }{\mleft\| \bs{\delta}
\mright\|^{3}}\,} \mright] \mright)\\
& = {\,\frac{1}{\sigma_e}\,} \mleft( \mb{I}_{2} - \varrho
{\,\frac{1}{\mleft\| \bs{\delta} \mright\|}\,} \mleft[ \mb{I}_{2} -
\bs{\delta}_0 \bs{\delta}_0^T \mright] \mright).
\end{align}

\subsection{Second Derivatives of Rotation Error}\label{sec_second_rot}

%
We compute the second derivatives of the rotation error from the generic
solutions in section \ref{sec_second_genrot}. For the first form we
obtain
\begin{align}
&\phantom{{}={}} \frac{\partial}{\partial \mb{u}}\,\mleft( \frac{\partial}{\partial
\mb{u}}\,g\mleft( \mb{p},\mb{p}' \mright) \mright)^T \nonumber \\
& = \gamma {\,\frac{1}{\sigma^{2}}\,} \bar{t}_1 {\,\frac{\mleft\| \mb{u}'
\mright\|}{\mleft\| \mb{u} \mright\|}\,} \mleft( \mb{I}_{2} - \mb{u}_0
\mb{u}_0^T \mright)
\end{align}
\parbox{\textwidth}{\rule{\textwidth}{0.4pt}}
\begin{align}
&\phantom{{}={}} \frac{\partial}{\partial \mb{u}'}\,\mleft( \frac{\partial}{\partial
\mb{u}}\,g\mleft( \mb{p},\mb{p}' \mright) \mright)^T \nonumber \\
& = \gamma {\,\frac{1}{\sigma^{2}}\,} \mleft( \bar{t}_1 \mb{u}_0 {{}
{\mb{u}'_0}}^T - \mb{Q}^T \mright)
\end{align}
\parbox{\textwidth}{\rule{\textwidth}{0.4pt}}
\begin{align}
&\phantom{{}={}} \frac{\partial}{\partial \mb{u}'}\,\mleft( \frac{\partial}{\partial
\mb{u}'}\,g\mleft( \mb{p},\mb{p}' \mright) \mright)^T \nonumber \\
& = \gamma {\,\frac{1}{\sigma^{2}}\,} \bar{t}_1 {\,\frac{\mleft\| \mb{u}
\mright\|}{\mleft\| \mb{u}' \mright\|}\,} \mleft( \mb{I}_{2} - {{}
{\mb{u}'_0}} {{} {\mb{u}'_0}}^T \mright).
\end{align}
For the second form we get
\begin{align}
&\phantom{{}={}} \frac{\partial}{\partial \mb{u}}\,\mleft( \frac{\partial}{\partial
\mb{u}}\,\bar{g}\mleft( \mb{p},\mb{p}' \mright) \mright)^T \nonumber \\
& = \gamma {\,\frac{1}{\sigma^{2}}\,} {\,\frac{1}{\mleft\| \mb{u}
\mright\|^{2}}\,} \mleft( \mb{Q}^T {{} {\mb{u}'_0}} \mb{u}_0^T + \mleft[
\mb{u}_0^T \mb{Q}^T {{} {\mb{u}'_0}} \mright] \mleft[ \mb{I}_{2} - 3
\mb{u}_0 \mb{u}_0^T \mright] + \mb{u}_0 {{} {\mb{u}'_0}}^T \mb{Q}
\mright)
\end{align}
\parbox{\textwidth}{\rule{\textwidth}{0.4pt}}
\begin{align}
&\phantom{{}={}} \frac{\partial}{\partial \mb{u}'}\,\mleft( \frac{\partial}{\partial
\mb{u}}\,\bar{g}\mleft( \mb{p},\mb{p}' \mright) \mright)^T \nonumber \\
& = - \gamma {\,\frac{1}{\sigma^{2}}\,} {\,\frac{1}{\mleft\| \mb{u}
\mright\| \mleft\| \mb{u}' \mright\| }\,} \mleft( \mb{I}_{2} - \mb{u}_0
\mb{u}_0^T \mright) \mb{Q}^T \mleft( \mb{I}_{2} - {{} {\mb{u}'_0}} {{}
{\mb{u}'_0}}^T \mright)
\end{align}
\parbox{\textwidth}{\rule{\textwidth}{0.4pt}}
\begin{align}
&\phantom{{}={}} \frac{\partial}{\partial \mb{u}'}\,\mleft( \frac{\partial}{\partial
\mb{u}'}\,\bar{g}\mleft( \mb{p},\mb{p}' \mright) \mright)^T \nonumber \\
& = \gamma {\,\frac{1}{\sigma^{2}}\,} {\,\frac{1}{\mleft\| \mb{u}'
\mright\|^{2}}\,} \mleft( \mb{Q} \mb{u}_0 {{} {\mb{u}'_0}}^T + \mleft[
{{} {\mb{u}'_0}}^T \mb{Q} \mb{u}_0 \mright] \mleft[ \mb{I}_{2} - 3 {{}
{\mb{u}'_0}} {{} {\mb{u}'_0}}^T \mright] + {{} {\mb{u}'_0}} \mb{u}_0^T
\mb{Q}^T \mright).
\end{align}

\subsection{Second Derivatives of Home-Vector Error}\label{sec_second_home}

\subsubsection{First Form}

%
We compute the second derivatives of the home-vector error for the first
form.

\textbf{For first derivative with respect to $\mb{x}$:}
\begin{align}
&\phantom{{}={}} \mleft( \frac{\partial}{\partial \mb{x}}\,h\mleft( \mb{p},\mb{p}'
\mright) \mright)^T \nonumber \\
& = \gamma {\,\frac{1}{\sigma_h^{2}}\,} \mleft(
{\,\frac{\mb{I}_{2}}{\mleft\| \bs{\delta} \mright\|}\,} -
{\,\frac{\bs{\delta} \bs{\delta}^T }{\mleft\| \bs{\delta}
\mright\|^{3}}\,} \mright) \bs{A} \mb{u}.
\end{align}
\textbf{Second derivatives:}
\begin{align}
&\phantom{{}={}} \frac{\partial}{\partial \mb{x}}\,\mleft( \frac{\partial}{\partial
\mb{x}}\,h\mleft( \mb{p},\mb{p}' \mright) \mright)^T \nonumber \\
& = - \frac{\partial}{\partial \bs{\delta}}\,\mleft(
\frac{\partial}{\partial \mb{x}}\,h\mleft( \mb{p},\mb{p}' \mright)
\mright)^T\\
& = - \frac{\partial}{\partial \bs{\delta}}\,\mleft( \gamma
{\,\frac{1}{\sigma_h^{2}}\,} \mleft[ {\,\frac{\mb{I}_{2}}{\mleft\|
\bs{\delta} \mright\|}\,} - {\,\frac{\bs{\delta} \bs{\delta}^T
}{\mleft\| \bs{\delta} \mright\|^{3}}\,} \mright] \bs{A} \mb{u} \mright)\\
& = - \frac{\partial}{\partial \bs{\delta}}\,\mleft( \gamma
{\,\frac{1}{\sigma_h^{2}}\,} \mleft[ {\,\frac{\bs{A} \mb{u} }{\mleft\|
\bs{\delta} \mright\|}\,} - {\,\frac{\bs{\delta} \bs{\delta}^T \bs{A}
\mb{u} }{\mleft\| \bs{\delta} \mright\|^{3}}\,} \mright] \mright)\\
& = - \gamma {\,\frac{1}{\sigma_h^{2}}\,} \mleft( \mleft[ - {\,\frac{\bs{A}
\mb{u} \bs{\delta}^T }{\mleft\| \bs{\delta} \mright\|^{3}}\,} \mright] -
\mleft[ {\,\frac{\mleft\{ \bs{\delta}^T \bs{A} \mb{u} \mright\}
\mb{I}_{2} }{\mleft\| \bs{\delta} \mright\|^{3}}\,} - 3
{\,\frac{\mleft\{ \bs{\delta}^T \bs{A} \mb{u} \mright\} \bs{\delta}
\bs{\delta}^T }{\mleft\| \bs{\delta} \mright\|^{5}}\,} +
{\,\frac{\bs{\delta} \mb{u}^T \bs{A}^T }{\mleft\| \bs{\delta}
\mright\|^{3}}\,} \mright] \mright)\\
& = \gamma {\,\frac{1}{\sigma_h^{2}}\,} {\,\frac{1}{\mleft\| \bs{\delta}
\mright\|^{2}}\,} \mleft( {\,\frac{\bs{A} \mb{u} \bs{\delta}^T
}{\mleft\| \bs{\delta} \mright\|}\,} + {\,\frac{\mleft[ \bs{\delta}^T
\bs{A} \mb{u} \mright] \mb{I}_{2} }{\mleft\| \bs{\delta} \mright\|}\,} -
3 {\,\frac{\mleft[ \bs{\delta}^T \bs{A} \mb{u} \mright] \bs{\delta}
\bs{\delta}^T }{\mleft\| \bs{\delta} \mright\|^{3}}\,} +
{\,\frac{\bs{\delta} \mb{u}^T \bs{A}^T }{\mleft\| \bs{\delta}
\mright\|}\,} \mright)\\
& = \gamma {\,\frac{1}{\sigma_h^{2}}\,} {\,\frac{1}{\mleft\| \bs{\delta}
\mright\|^{2}}\,} \mleft( \bs{A} \mb{u} \bs{\delta}_0^T + \mleft[
\bs{\delta}_0^T \bs{A} \mb{u} \mright] \mb{I}_{2} - 3 \mleft[
\bs{\delta}_0^T \bs{A} \mb{u} \mright] \bs{\delta}_0 \bs{\delta}_0^T +
\bs{\delta}_0 \mb{u}^T \bs{A}^T \mright)\\
& = \gamma {\,\frac{1}{\sigma_h^{2}}\,} {\,\frac{1}{\mleft\| \bs{\delta}
\mright\|^{2}}\,} \mleft( \bs{A} \mb{u} \bs{\delta}_0^T + \mleft[
\bs{\delta}_0^T \bs{A} \mb{u} \mright] \mleft[ \mb{I}_{2} - 3
\bs{\delta}_0 \bs{\delta}_0^T \mright] + \bs{\delta}_0 \mb{u}^T \bs{A}^T
\mright)
\end{align}
\parbox{\textwidth}{\rule{\textwidth}{0.4pt}}
\begin{align}
&\phantom{{}={}} \frac{\partial}{\partial \mb{x}'}\,\mleft( \frac{\partial}{\partial
\mb{x}}\,h\mleft( \mb{p},\mb{p}' \mright) \mright)^T \nonumber \\
& = \frac{\partial}{\partial \bs{\delta}}\,\mleft( \frac{\partial}{\partial
\mb{x}}\,h\mleft( \mb{p},\mb{p}' \mright) \mright)^T\\
& = - \gamma {\,\frac{1}{\sigma_h^{2}}\,} {\,\frac{1}{\mleft\| \bs{\delta}
\mright\|^{2}}\,} \mleft( \bs{A} \mb{u} \bs{\delta}_0^T + \mleft[
\bs{\delta}_0^T \bs{A} \mb{u} \mright] \mleft[ \mb{I}_{2} - 3
\bs{\delta}_0 \bs{\delta}_0^T \mright] + \bs{\delta}_0 \mb{u}^T \bs{A}^T
\mright)
\end{align}
\parbox{\textwidth}{\rule{\textwidth}{0.4pt}}
\begin{align}
&\phantom{{}={}} \frac{\partial}{\partial \mb{u}}\,\mleft( \frac{\partial}{\partial
\mb{x}}\,h\mleft( \mb{p},\mb{p}' \mright) \mright)^T \nonumber \\
& = \gamma {\,\frac{1}{\sigma_h^{2}}\,} \mleft(
{\,\frac{\mb{I}_{2}}{\mleft\| \bs{\delta} \mright\|}\,} -
{\,\frac{\bs{\delta} \bs{\delta}^T }{\mleft\| \bs{\delta}
\mright\|^{3}}\,} \mright) \bs{A}\\
& = \gamma {\,\frac{1}{\sigma_h^{2}}\,} {\,\frac{1}{\mleft\| \bs{\delta}
\mright\|}\,} \mleft( \mb{I}_{2} - \bs{\delta}_0 \bs{\delta}_0^T
\mright) \bs{A}.
\end{align}
\textbf{For first derivative with respect to $\mb{x}'$:}
\begin{align}
&\phantom{{}={}} \mleft( \frac{\partial}{\partial \mb{x}'}\,h\mleft( \mb{p},\mb{p}'
\mright) \mright)^T \nonumber \\
& = - \gamma {\,\frac{1}{\sigma_h^{2}}\,} \mleft(
{\,\frac{\mb{I}_{2}}{\mleft\| \bs{\delta} \mright\|}\,} -
{\,\frac{\bs{\delta} \bs{\delta}^T }{\mleft\| \bs{\delta}
\mright\|^{3}}\,} \mright) \bs{A} \mb{u}.
\end{align}
\textbf{Second derivatives:}
\begin{align}
&\phantom{{}={}} \frac{\partial}{\partial \mb{x}'}\,\mleft( \frac{\partial}{\partial
\mb{x}'}\,h\mleft( \mb{p},\mb{p}' \mright) \mright)^T \nonumber \\
& = \frac{\partial}{\partial \bs{\delta}}\,\mleft( \frac{\partial}{\partial
\mb{x}'}\,h\mleft( \mb{p},\mb{p}' \mright) \mright)^T\\
& = \gamma {\,\frac{1}{\sigma_h^{2}}\,} {\,\frac{1}{\mleft\| \bs{\delta}
\mright\|^{2}}\,} \mleft( \bs{A} \mb{u} \bs{\delta}_0^T + \mleft[
\bs{\delta}_0^T \bs{A} \mb{u} \mright] \mleft[ \mb{I}_{2} - 3
\bs{\delta}_0 \bs{\delta}_0^T \mright] + \bs{\delta}_0 \mb{u}^T \bs{A}^T
\mright)
\end{align}
\parbox{\textwidth}{\rule{\textwidth}{0.4pt}}
\begin{align}
&\phantom{{}={}} \frac{\partial}{\partial \mb{u}}\,\mleft( \frac{\partial}{\partial
\mb{x}'}\,h\mleft( \mb{p},\mb{p}' \mright) \mright)^T \nonumber \\
& = - \gamma {\,\frac{1}{\sigma_h^{2}}\,} {\,\frac{1}{\mleft\| \bs{\delta}
\mright\|}\,} \mleft( \mb{I}_{2} - \bs{\delta}_0 \bs{\delta}_0^T
\mright) \bs{A}.
\end{align}
\textbf{For first derivative with respect to $\mb{u}$:}
\begin{align}
&\phantom{{}={}} \mleft( \frac{\partial}{\partial \mb{u}}\,h\mleft( \mb{p},\mb{p}'
\mright) \mright)^T \nonumber \\
& = \gamma {\,\frac{1}{\sigma_h^{2}}\,} \mleft( \bar{t}_1
{\,\frac{\mb{u}}{\mleft\| \mb{u} \mright\|}\,} - \bs{A}^T \bs{\delta}_0
\mright).
\end{align}
\textbf{Second derivative:}
\begin{align}
&\phantom{{}={}} \frac{\partial}{\partial \mb{u}}\,\mleft( \frac{\partial}{\partial
\mb{u}}\,h\mleft( \mb{p},\mb{p}' \mright) \mright)^T \nonumber \\
& = \gamma {\,\frac{1}{\sigma_h^{2}}\,} \bar{t}_1 {\,\frac{1}{\mleft\|
\mb{u} \mright\|}\,} \mleft( \mb{I}_{2} - \mb{u}_0 \mb{u}_0^T \mright).
\end{align}

\subsubsection{Second Form}

%
We compute the second derivatives of the home-vector error for the
second form. See the derivation for the first form for some intermediate
steps.

\textbf{For first derivative with respect to $\mb{x}$:}
\begin{align}
&\phantom{{}={}} \mleft( \frac{\partial}{\partial \mb{x}}\,\bar{h}\mleft( \mb{p},\mb{p}'
\mright) \mright)^T \nonumber \\
& = \gamma {\,\frac{1}{\sigma_h^{2}}\,} \mleft(
{\,\frac{\mb{I}_{2}}{\mleft\| \bs{\delta} \mright\|}\,} -
{\,\frac{\bs{\delta} \bs{\delta}^T }{\mleft\| \bs{\delta}
\mright\|^{3}}\,} \mright) \bs{A} \mb{u}_0.
\end{align}
\textbf{Second derivatives:}
\begin{align}
&\phantom{{}={}} \frac{\partial}{\partial \mb{x}}\,\mleft( \frac{\partial}{\partial
\mb{x}}\,\bar{h}\mleft( \mb{p},\mb{p}' \mright) \mright)^T \nonumber \\
& = \gamma {\,\frac{1}{\sigma_h^{2}}\,} {\,\frac{1}{\mleft\| \bs{\delta}
\mright\|^{2}}\,} \mleft( \bs{A} \mb{u}_0 \bs{\delta}_0^T + \mleft[
\bs{\delta}_0^T \bs{A} \mb{u}_0 \mright] \mleft[ \mb{I}_{2} - 3
\bs{\delta}_0 \bs{\delta}_0^T \mright] + \bs{\delta}_0 \mb{u}_0^T
\bs{A}^T \mright)
\end{align}
\parbox{\textwidth}{\rule{\textwidth}{0.4pt}}
\begin{align}
&\phantom{{}={}} \frac{\partial}{\partial \mb{x}'}\,\mleft( \frac{\partial}{\partial
\mb{x}}\,\bar{h}\mleft( \mb{p},\mb{p}' \mright) \mright)^T \nonumber \\
& = - \gamma {\,\frac{1}{\sigma_h^{2}}\,} {\,\frac{1}{\mleft\| \bs{\delta}
\mright\|^{2}}\,} \mleft( \bs{A} \mb{u}_0 \bs{\delta}_0^T + \mleft[
\bs{\delta}_0^T \bs{A} \mb{u}_0 \mright] \mleft[ \mb{I}_{2} - 3
\bs{\delta}_0 \bs{\delta}_0^T \mright] + \bs{\delta}_0 \mb{u}_0^T
\bs{A}^T \mright)
\end{align}
\parbox{\textwidth}{\rule{\textwidth}{0.4pt}}
\begin{align}
&\phantom{{}={}} \frac{\partial}{\partial \mb{u}}\,\mleft( \frac{\partial}{\partial
\mb{x}}\,\bar{h}\mleft( \mb{p},\mb{p}' \mright) \mright)^T \nonumber \\
& = \gamma {\,\frac{1}{\sigma_h^{2}}\,} \mleft(
{\,\frac{\mb{I}_{2}}{\mleft\| \bs{\delta} \mright\|}\,} -
{\,\frac{\bs{\delta} \bs{\delta}^T }{\mleft\| \bs{\delta}
\mright\|^{3}}\,} \mright) \bs{A}\\
& = \gamma {\,\frac{1}{\sigma_h^{2}}\,} {\,\frac{1}{\mleft\| \bs{\delta}
\mright\| \mleft\| \mb{u} \mright\| }\,} \mleft( \mb{I}_{2} -
\bs{\delta}_0 \bs{\delta}_0^T \mright) \bs{A} \mleft( \mb{I}_{2} -
\mb{u}_0 \mb{u}_0^T \mright).
\end{align}
\textbf{For first derivative with respect to $\mb{x}'$:}
\begin{align}
&\phantom{{}={}} \mleft( \frac{\partial}{\partial \mb{x}'}\,\bar{h}\mleft( \mb{p},\mb{p}'
\mright) \mright)^T \nonumber \\
& = - \gamma {\,\frac{1}{\sigma_h^{2}}\,} \mleft(
{\,\frac{\mb{I}_{2}}{\mleft\| \bs{\delta} \mright\|}\,} -
{\,\frac{\bs{\delta} \bs{\delta}^T }{\mleft\| \bs{\delta}
\mright\|^{3}}\,} \mright) \bs{A} \mb{u}_0.
\end{align}
\textbf{Second derivatives:}
\begin{align}
&\phantom{{}={}} \frac{\partial}{\partial \mb{x}'}\,\mleft( \frac{\partial}{\partial
\mb{x}'}\,\bar{h}\mleft( \mb{p},\mb{p}' \mright) \mright)^T \nonumber \\
& = \gamma {\,\frac{1}{\sigma_h^{2}}\,} {\,\frac{1}{\mleft\| \bs{\delta}
\mright\|^{2}}\,} \mleft( \bs{A} \mb{u}_0 \bs{\delta}_0^T + \mleft[
\bs{\delta}_0^T \bs{A} \mb{u}_0 \mright] \mleft[ \mb{I}_{2} - 3
\bs{\delta}_0 \bs{\delta}_0^T \mright] + \bs{\delta}_0 \mb{u}_0^T
\bs{A}^T \mright)
\end{align}
\parbox{\textwidth}{\rule{\textwidth}{0.4pt}}
\begin{align}
&\phantom{{}={}} \frac{\partial}{\partial \mb{u}}\,\mleft( \frac{\partial}{\partial
\mb{x}'}\,\bar{h}\mleft( \mb{p},\mb{p}' \mright) \mright)^T \nonumber \\
& = - \gamma {\,\frac{1}{\sigma_h^{2}}\,} {\,\frac{1}{\mleft\| \bs{\delta}
\mright\| \mleft\| \mb{u} \mright\| }\,} \mleft( \mb{I}_{2} -
\bs{\delta}_0 \bs{\delta}_0^T \mright) \bs{A} \mleft( \mb{I}_{2} -
\mb{u}_0 \mb{u}_0^T \mright).
\end{align}
\textbf{For first derivative with respect to $\mb{u}$:}
\begin{align}
&\phantom{{}={}} \mleft( \frac{\partial}{\partial \mb{u}}\,\bar{h}\mleft( \mb{p},\mb{p}'
\mright) \mright)^T \nonumber \\
& = - \gamma {\,\frac{1}{\sigma_h^{2}}\,} \mleft(
{\,\frac{\mb{I}_{2}}{\mleft\| \mb{u} \mright\|}\,} - {\,\frac{\mb{u}
\mb{u}^T }{\mleft\| \mb{u} \mright\|^{3}}\,} \mright) \bs{A}^T
\bs{\delta}_0.
\end{align}
\textbf{Second derivative:}
\begin{align}
&\phantom{{}={}} \frac{\partial}{\partial \mb{u}}\,\mleft( \frac{\partial}{\partial
\mb{u}}\,\bar{h}\mleft( \mb{p},\mb{p}' \mright) \mright)^T \nonumber \\
& = \gamma {\,\frac{1}{\sigma_h^{2}}\,} {\,\frac{1}{\mleft\| \mb{u}
\mright\|^{2}}\,} \mleft( \bs{A}^T \bs{\delta}_0 \mb{u}_0^T + \mleft[
\mb{u}_0^T \bs{A}^T \bs{\delta}_0 \mright] \mleft[ \mb{I}_{2} - 3
\mb{u}_0 \mb{u}_0^T \mright] + \mb{u}_0 \bs{\delta}_0^T \bs{A} \mright).
\end{align}

\subsection{Second Derivatives of Compass Error}\label{sec_second_compass}

%
We compute the second derivatives of the compass error from the generic
solutions in section \ref{sec_second_genrot}. For the first form we
obtain
\begin{align}
&\phantom{{}={}} \frac{\partial}{\partial \mb{u}}\,\mleft( \frac{\partial}{\partial
\mb{u}}\,g\mleft( \mb{p},\mb{p}' \mright) \mright)^T \nonumber \\
& = \gamma {\,\frac{1}{\sigma_c^{2}}\,} \bar{t}_1 {\,\frac{\mleft\| \mb{u}'
\mright\|}{\mleft\| \mb{u} \mright\|}\,} \mleft( \mb{I}_{2} - \mb{u}_0
\mb{u}_0^T \mright)
\end{align}
\parbox{\textwidth}{\rule{\textwidth}{0.4pt}}
\begin{align}
&\phantom{{}={}} \frac{\partial}{\partial \mb{u}'}\,\mleft( \frac{\partial}{\partial
\mb{u}}\,g\mleft( \mb{p},\mb{p}' \mright) \mright)^T \nonumber \\
& = \gamma {\,\frac{1}{\sigma_c^{2}}\,} \mleft( \bar{t}_1 \mb{u}_0 {{}
{\mb{u}'_0}}^T - \bs{\Psi}^T \mright)
\end{align}
\parbox{\textwidth}{\rule{\textwidth}{0.4pt}}
\begin{align}
&\phantom{{}={}} \frac{\partial}{\partial \mb{u}'}\,\mleft( \frac{\partial}{\partial
\mb{u}'}\,g\mleft( \mb{p},\mb{p}' \mright) \mright)^T \nonumber \\
& = \gamma {\,\frac{1}{\sigma_c^{2}}\,} \bar{t}_1 {\,\frac{\mleft\| \mb{u}
\mright\|}{\mleft\| \mb{u}' \mright\|}\,} \mleft( \mb{I}_{2} - {{}
{\mb{u}'_0}} {{} {\mb{u}'_0}}^T \mright).
\end{align}
For the second form we get
\begin{align}
&\phantom{{}={}} \frac{\partial}{\partial \mb{u}}\,\mleft( \frac{\partial}{\partial
\mb{u}}\,\bar{g}\mleft( \mb{p},\mb{p}' \mright) \mright)^T \nonumber \\
& = \gamma {\,\frac{1}{\sigma_c^{2}}\,} {\,\frac{1}{\mleft\| \mb{u}
\mright\|^{2}}\,} \mleft( \bs{\Psi}^T {{} {\mb{u}'_0}} \mb{u}_0^T +
\mleft[ \mb{u}_0^T \bs{\Psi}^T {{} {\mb{u}'_0}} \mright] \mleft[
\mb{I}_{2} - 3 \mb{u}_0 \mb{u}_0^T \mright] + \mb{u}_0 {{}
{\mb{u}'_0}}^T \bs{\Psi} \mright)
\end{align}
\parbox{\textwidth}{\rule{\textwidth}{0.4pt}}
\begin{align}
&\phantom{{}={}} \frac{\partial}{\partial \mb{u}'}\,\mleft( \frac{\partial}{\partial
\mb{u}}\,\bar{g}\mleft( \mb{p},\mb{p}' \mright) \mright)^T \nonumber \\
& = - \gamma {\,\frac{1}{\sigma_c^{2}}\,} {\,\frac{1}{\mleft\| \mb{u}
\mright\| \mleft\| \mb{u}' \mright\| }\,} \mleft( \mb{I}_{2} - \mb{u}_0
\mb{u}_0^T \mright) \bs{\Psi}^T \mleft( \mb{I}_{2} - {{} {\mb{u}'_0}}
{{} {\mb{u}'_0}}^T \mright)
\end{align}
\parbox{\textwidth}{\rule{\textwidth}{0.4pt}}
\begin{align}
&\phantom{{}={}} \frac{\partial}{\partial \mb{u}'}\,\mleft( \frac{\partial}{\partial
\mb{u}'}\,\bar{g}\mleft( \mb{p},\mb{p}' \mright) \mright)^T \nonumber \\
& = \gamma {\,\frac{1}{\sigma_c^{2}}\,} {\,\frac{1}{\mleft\| \mb{u}'
\mright\|^{2}}\,} \mleft( \bs{\Psi} \mb{u}_0 {{} {\mb{u}'_0}}^T +
\mleft[ {{} {\mb{u}'_0}}^T \bs{\Psi} \mb{u}_0 \mright] \mleft[
\mb{I}_{2} - 3 {{} {\mb{u}'_0}} {{} {\mb{u}'_0}}^T \mright] + {{}
{\mb{u}'_0}} \mb{u}_0^T \bs{\Psi}^T \mright).
\end{align}

\section{Lagrange-Newton Descent}\label{sec_lagrange_newton}

\subsection{State Vector}\label{sec_state}

%
The state vector for the Lagrange-Newton descent comprises all poses,
split into the position vectors ${\mb{x}}_{i}$ and the orientation
vectors ${\mb{u}}_{i}$, and all Lagrange parameters ${\lambda}_{i}$.
These terms contain all indices $i$ except $1$, since pose $1$ is
assumed to be fixed.

\subsection{Constraint Terms}\label{sec_constraints}

%
For each orientation vector ${\mb{u}}_{i}$, we have a constraint term
\begin{align}
w\mleft( {\lambda}_{i},{\mb{u}}_{i} \mright) & = {\lambda}_{i} l\mleft( {\mb{u}}_{i} \mright) = \frac{1}{2}\, {\lambda}_{i} \mleft( {\mb{u}}_{i}^T {\mb{u}}_{i} - 1
\mright),
\end{align}
its first derivatives
\begin{align}
\frac{\partial}{\partial {\mb{u}}_{i}}\,w\mleft(
{\lambda}_{i},{\mb{u}}_{i} \mright) & = \frac{\partial}{\partial {\mb{u}}_{i}}\,\mleft( \frac{1}{2}\,
{\lambda}_{i} \mleft[ {\mb{u}}_{i}^T {\mb{u}}_{i} - 1 \mright] \mright) = {\lambda}_{i} {\mb{u}}_{i}^T\\
\frac{\partial}{\partial {\lambda}_{i}}\,w\mleft(
{\lambda}_{i},{\mb{u}}_{i} \mright) & = \frac{\partial}{\partial {\lambda}_{i}}\,\mleft( \frac{1}{2}\,
{\lambda}_{i} \mleft[ {\mb{u}}_{i}^T {\mb{u}}_{i} - 1 \mright] \mright) = \frac{1}{2}\, \mleft( {\mb{u}}_{i}^T {\mb{u}}_{i} - 1 \mright) = l\mleft( {\mb{u}}_{i} \mright),
\end{align}
and its second derivatives
\begin{align}
\frac{\partial}{\partial {\mb{u}}_{i}}\,\mleft( \frac{\partial}{\partial
{\mb{u}}_{i}}\,w\mleft( {\lambda}_{i},{\mb{u}}_{i} \mright) \mright)^T
&=
{\lambda}_{i} \mb{I}_{2}\\
\frac{\partial}{\partial {\lambda}_{i}}\,\mleft(
\frac{\partial}{\partial {\mb{u}}_{i}}\,w\mleft(
{\lambda}_{i},{\mb{u}}_{i} \mright) \mright)^T
&=
{\mb{u}}_{i}\\
\frac{\partial}{\partial {\mb{u}}_{i}}\,\mleft( \frac{\partial}{\partial
{\lambda}_{i}}\,w\mleft( {\lambda}_{i},{\mb{u}}_{i} \mright) \mright)^T
&=
{\mb{u}}_{i}^T\\
\frac{\partial}{\partial {\lambda}_{i}}\,\mleft(
\frac{\partial}{\partial {\lambda}_{i}}\,w\mleft(
{\lambda}_{i},{\mb{u}}_{i} \mright) \mright)^T
&=
0.
\end{align}
Since we compute the derivatives with respect to an entire pose
$\mb{p}$, we need to insert zero terms corresponding to the component
${\mb{x}}_{i}$ (see below).

\subsection{Total Cost Function}\label{sec_total_cost}

%
The total cost function is the ``Lagrangian'' which is formed for all
individual cost terms for the five types of measurements and the
constraint terms. Assuming that all of them are used, the sum of all
five cost terms is:
\begin{align}
\label{eq_F}
F
&=
\sum\limits_{m = 1}^{M_o}f\mleft(
{\mb{p}}_{i_1({m})},{\mb{p}}_{i_2({m})},{\mb{T}}_{m},{\mb{r}}_{m}
\mright) + \sum\limits_{m = 1}^{M_o}e\mleft(
{\mb{p}}_{i_1({m})},{\mb{p}}_{i_2({m})},{\sigma_e}_{m},{\mb{r}}_{m}
\mright)
\\&\phantom{{}={}}
\nonumber
{} \! + \sum\limits_{m = 1}^{M_o}g\mleft(
{\mb{p}}_{i_1({m})},{\mb{p}}_{i_2({m})},{\mb{Q}}_{m},{\sigma}_{m}
\mright)
\\&\phantom{{}={}}
\nonumber
{} \! + \sum\limits_{m = 1}^{M_h}h\mleft(
{\mb{p}}_{i_1({m})},{\mb{p}}_{i_2({m})},{\bs{A}}_{m},{\sigma_h}_{m}
\mright) + \sum\limits_{m = 1}^{M_h}c\mleft(
{\mb{p}}_{i_1({m})},{\mb{p}}_{i_2({m})},{\bs{\Psi}}_{m},{\sigma_c}_{m}
\mright),
\end{align}
where the mappings $i_1$ and $i_2$ determine the indices of the two
poses used for measurement $m$, and the total cost function
(``Lagrangian'') including the constraint terms is
\begin{align}
L
&=
F + \sum\limits_{i = 2}^{N}w\mleft( {\lambda}_{i},{\mb{u}}_{i} \mright).
\end{align}

\subsection{Gradient of Total Cost Function}\label{sec_total_gradient}

%
The gradient of the cost functions is shown for the example of the
translation error $f\mleft( \mb{p},\mb{p}' \mright)$; the other four
cost functions are handled in the same way. Note that we have two
gradient terms, one for the pose $\mb{p}$ appearing as the first
argument in the cost function, the other for the pose $\mb{p}'$
appearing as the second argument:
\begin{align}
\frac{\partial}{\partial \mb{p}}\,f\mleft( \mb{p},\mb{p}' \mright)
&=
\begin{pmatrix}\frac{\partial}{\partial \mb{x}}\,f\mleft( \mb{p},\mb{p}'
\mright) & \frac{\partial}{\partial \mb{u}}\,f\mleft( \mb{p},\mb{p}'
\mright)\end{pmatrix}\\
\frac{\partial}{\partial \mb{p}'}\,f\mleft( \mb{p},\mb{p}' \mright)
&=
\begin{pmatrix}\frac{\partial}{\partial \mb{x}'}\,f\mleft(
\mb{p},\mb{p}' \mright) & \frac{\partial}{\partial \mb{u}'}\,f\mleft(
\mb{p},\mb{p}' \mright)\end{pmatrix}.
\end{align}
Note that we always have $\mb{p} \ne \mb{p}'$.

The gradient of the total cost function with respect to a single pose
${\mb{p}}_{k}$ (with $k \neq 1$) is (with the four other cost functions
omitted)
\begin{align}
&\phantom{{}={}} \frac{\partial}{\partial {\mb{p}}_{k}}\,L \nonumber \\
& = \sum\limits_{m = 1}^{M_o}\frac{\partial}{\partial
{\mb{p}}_{k}}\,f\mleft(
{\mb{p}}_{i_1({m})},{\mb{p}}_{i_2({m})},{\mb{T}}_{m},{\mb{r}}_{m}
\mright) + \ldots + \sum\limits_{i = 2}^{N}\frac{\partial}{\partial
{\mb{p}}_{k}}\,w\mleft( {\lambda}_{i},{\mb{u}}_{i} \mright)\\
& = \sum\limits_{m = 1}^{M_o}\mleft( {\mathfrak{d}}_{k,i_1({m})}
\mleft.{\mleft[ \frac{\partial}{\partial \mb{p}}\,f\mleft(
\mb{p},\mb{p}' \mright)
\mright]}\mright|_{{\mb{p}}_{i_1({m})},{\mb{p}}_{i_2({m})},{\mb{T}}_{m},{\mb{r}}_{m}}
\vphantom{{\mathfrak{d}}_{k,i_1({m})} \mleft.{\mleft[
\frac{\partial}{\partial \mb{p}}\,f\mleft( \mb{p},\mb{p}' \mright)
\mright]}\mright|_{{\mb{p}}_{i_1({m})},{\mb{p}}_{i_2({m})},{\mb{T}}_{m},{\mb{r}}_{m}}
+ {\mathfrak{d}}_{k,i_2({m})} \mleft.{\mleft[ \frac{\partial}{\partial
\mb{p}'}\,f\mleft( \mb{p},\mb{p}' \mright)
\mright]}\mright|_{{\mb{p}}_{i_1({m})},{\mb{p}}_{i_2({m})},{\mb{T}}_{m},{\mb{r}}_{m}}
} \mright.
\\&\phantom{{}={}}
\nonumber
{} \! + \mleft. \vphantom{{\mathfrak{d}}_{k,i_1({m})} \mleft.{\mleft[
\frac{\partial}{\partial \mb{p}}\,f\mleft( \mb{p},\mb{p}' \mright)
\mright]}\mright|_{{\mb{p}}_{i_1({m})},{\mb{p}}_{i_2({m})},{\mb{T}}_{m},{\mb{r}}_{m}}
+ {\mathfrak{d}}_{k,i_2({m})} \mleft.{\mleft[ \frac{\partial}{\partial
\mb{p}'}\,f\mleft( \mb{p},\mb{p}' \mright)
\mright]}\mright|_{{\mb{p}}_{i_1({m})},{\mb{p}}_{i_2({m})},{\mb{T}}_{m},{\mb{r}}_{m}}
} {\mathfrak{d}}_{k,i_2({m})} \mleft.{\mleft[ \frac{\partial}{\partial
\mb{p}'}\,f\mleft( \mb{p},\mb{p}' \mright)
\mright]}\mright|_{{\mb{p}}_{i_1({m})},{\mb{p}}_{i_2({m})},{\mb{T}}_{m},{\mb{r}}_{m}}
\mright)
\\&\phantom{{}={}}
\nonumber
{} \! + \ldots
\\&\phantom{{}={}}
\nonumber
{} \! + \begin{pmatrix}\mb{0}_{2}^T & {\lambda}_{k} {\mb{u}}_{k}^T
\end{pmatrix}
\end{align}
and with respect to the Lagrange multiplier ${\lambda}_{k}$ (with $k
\neq 1$)
\begin{align}
&\phantom{{}={}} \frac{\partial}{\partial {\lambda}_{k}}\,L \nonumber \\
& = \frac{1}{2}\, \mleft( {\mb{u}}_{k}^T {\mb{u}}_{k} - 1 \mright).
\end{align}

\subsection{Hessian of Total Cost Function}\label{sec_total_hessian}

%
We assemble the Hessian from blocks, where each block relates to
combinations of pose and corresponding Lagrange multiplier (in both
dimensions of the matrix). Also the Hessian blocks are only shown for
the example of the translation error $f\mleft( \mb{p},\mb{p}' \mright)$.
The total Hessian block is the sum of all Hessian blocks for the
different error functions. For the translation error (and analogous for
the four other error functions), we have four different versions of the
second derivatives of the translation error, since the two first
derivatives both depend on $\mb{p}$ and $\mb{p}'$, and we need second
derivatives for both arguments:
\begin{align}
\frac{\partial}{\partial \mb{p}}\,\mleft( \frac{\partial}{\partial
\mb{p}}\,f\mleft( \mb{p},\mb{p}' \mright) \mright)^T
&=
\begin{pmatrix}\frac{\partial}{\partial \mb{x}}\,\mleft(
\frac{\partial}{\partial \mb{x}}\,f\mleft( \mb{p},\mb{p}' \mright)
\mright)^T & \frac{\partial}{\partial \mb{u}}\,\mleft(
\frac{\partial}{\partial \mb{x}}\,f\mleft( \mb{p},\mb{p}' \mright)
\mright)^T\\\frac{\partial}{\partial \mb{x}}\,\mleft(
\frac{\partial}{\partial \mb{u}}\,f\mleft( \mb{p},\mb{p}' \mright)
\mright)^T & \frac{\partial}{\partial \mb{u}}\,\mleft(
\frac{\partial}{\partial \mb{u}}\,f\mleft( \mb{p},\mb{p}' \mright)
\mright)^T\end{pmatrix}\\
\frac{\partial}{\partial \mb{p}}\,\mleft( \frac{\partial}{\partial
\mb{p}'}\,f\mleft( \mb{p},\mb{p}' \mright) \mright)^T
&=
\begin{pmatrix}\frac{\partial}{\partial \mb{x}}\,\mleft(
\frac{\partial}{\partial \mb{x}'}\,f\mleft( \mb{p},\mb{p}' \mright)
\mright)^T & \frac{\partial}{\partial \mb{u}}\,\mleft(
\frac{\partial}{\partial \mb{x}'}\,f\mleft( \mb{p},\mb{p}' \mright)
\mright)^T\\\frac{\partial}{\partial \mb{x}}\,\mleft(
\frac{\partial}{\partial \mb{u}'}\,f\mleft( \mb{p},\mb{p}' \mright)
\mright)^T & \frac{\partial}{\partial \mb{u}}\,\mleft(
\frac{\partial}{\partial \mb{u}'}\,f\mleft( \mb{p},\mb{p}' \mright)
\mright)^T\end{pmatrix}\\
\frac{\partial}{\partial \mb{p}'}\,\mleft( \frac{\partial}{\partial
\mb{p}}\,f\mleft( \mb{p},\mb{p}' \mright) \mright)^T
&=
\begin{pmatrix}\frac{\partial}{\partial \mb{x}'}\,\mleft(
\frac{\partial}{\partial \mb{x}}\,f\mleft( \mb{p},\mb{p}' \mright)
\mright)^T & \frac{\partial}{\partial \mb{u}'}\,\mleft(
\frac{\partial}{\partial \mb{x}}\,f\mleft( \mb{p},\mb{p}' \mright)
\mright)^T\\\frac{\partial}{\partial \mb{x}'}\,\mleft(
\frac{\partial}{\partial \mb{u}}\,f\mleft( \mb{p},\mb{p}' \mright)
\mright)^T & \frac{\partial}{\partial \mb{u}'}\,\mleft(
\frac{\partial}{\partial \mb{u}}\,f\mleft( \mb{p},\mb{p}' \mright)
\mright)^T\end{pmatrix}\\
\frac{\partial}{\partial \mb{p}'}\,\mleft( \frac{\partial}{\partial
\mb{p}'}\,f\mleft( \mb{p},\mb{p}' \mright) \mright)^T
&=
\begin{pmatrix}\frac{\partial}{\partial \mb{x}'}\,\mleft(
\frac{\partial}{\partial \mb{x}'}\,f\mleft( \mb{p},\mb{p}' \mright)
\mright)^T & \frac{\partial}{\partial \mb{u}'}\,\mleft(
\frac{\partial}{\partial \mb{x}'}\,f\mleft( \mb{p},\mb{p}' \mright)
\mright)^T\\\frac{\partial}{\partial \mb{x}'}\,\mleft(
\frac{\partial}{\partial \mb{u}'}\,f\mleft( \mb{p},\mb{p}' \mright)
\mright)^T & \frac{\partial}{\partial \mb{u}'}\,\mleft(
\frac{\partial}{\partial \mb{u}'}\,f\mleft( \mb{p},\mb{p}' \mright)
\mright)^T\end{pmatrix}.
\end{align}
The Hessian block for the total cost function (again only the terms for
the translation error are shown) is determined from the four terms which
together describe a block of size $5\times 5$ at row position $k$ and
column position $l$. The first term is
\begin{align}
&\phantom{{}={}} \frac{\partial}{\partial {\mb{p}}_{l}}\,\mleft( \frac{\partial}{\partial
{\mb{p}}_{k}}\,L \mright)^T \nonumber \\
& = \sum\limits_{m = 1}^{M_o}\frac{\partial}{\partial {\mb{p}}_{l}}\,\mleft(
\frac{\partial}{\partial {\mb{p}}_{k}}\,f\mleft(
{\mb{p}}_{i_1({m})},{\mb{p}}_{i_2({m})},{\mb{T}}_{m},{\mb{r}}_{m}
\mright) \mright)^T + \ldots + \sum\limits_{i =
2}^{N}\frac{\partial}{\partial {\mb{p}}_{l}}\,\mleft(
\frac{\partial}{\partial {\mb{p}}_{k}}\,w\mleft(
{\lambda}_{i},{\mb{u}}_{i} \mright) \mright)^T\\
& = \sum\limits_{m = 1}^{M_o}\mleft( {\mathfrak{d}}_{l,i_1({m})}
{\mathfrak{d}}_{k,i_1({m})} \mleft.{\mleft[ \frac{\partial}{\partial
\mb{p}}\,\mleft\{ \frac{\partial}{\partial \mb{p}}\,f\mleft(
\mb{p},\mb{p}' \mright) \mright\}^T
\mright]}\mright|_{{\mb{p}}_{i_1({m})},{\mb{p}}_{i_2({m})},{\mb{T}}_{m},{\mb{r}}_{m}}
\vphantom{{\mathfrak{d}}_{l,i_1({m})} {\mathfrak{d}}_{k,i_1({m})}
\mleft.{\mleft[ \frac{\partial}{\partial \mb{p}}\,\mleft\{
\frac{\partial}{\partial \mb{p}}\,f\mleft( \mb{p},\mb{p}' \mright)
\mright\}^T
\mright]}\mright|_{{\mb{p}}_{i_1({m})},{\mb{p}}_{i_2({m})},{\mb{T}}_{m},{\mb{r}}_{m}}
+ {\mathfrak{d}}_{l,i_1({m})} {\mathfrak{d}}_{k,i_2({m})}
\mleft.{\mleft[ \frac{\partial}{\partial \mb{p}}\,\mleft\{
\frac{\partial}{\partial \mb{p}'}\,f\mleft( \mb{p},\mb{p}' \mright)
\mright\}^T
\mright]}\mright|_{{\mb{p}}_{i_1({m})},{\mb{p}}_{i_2({m})},{\mb{T}}_{m},{\mb{r}}_{m}}
+ {\mathfrak{d}}_{l,i_2({m})} {\mathfrak{d}}_{k,i_1({m})}
\mleft.{\mleft[ \frac{\partial}{\partial \mb{p}'}\,\mleft\{
\frac{\partial}{\partial \mb{p}}\,f\mleft( \mb{p},\mb{p}' \mright)
\mright\}^T
\mright]}\mright|_{{\mb{p}}_{i_1({m})},{\mb{p}}_{i_2({m})},{\mb{T}}_{m},{\mb{r}}_{m}}
+ {\mathfrak{d}}_{l,i_2({m})} {\mathfrak{d}}_{k,i_2({m})}
\mleft.{\mleft[ \frac{\partial}{\partial \mb{p}'}\,\mleft\{
\frac{\partial}{\partial \mb{p}'}\,f\mleft( \mb{p},\mb{p}' \mright)
\mright\}^T
\mright]}\mright|_{{\mb{p}}_{i_1({m})},{\mb{p}}_{i_2({m})},{\mb{T}}_{m},{\mb{r}}_{m}}
} \mright.
\\&\phantom{{}={}}
\nonumber
{} \! + \mleft. {\mathfrak{d}}_{l,i_1({m})} {\mathfrak{d}}_{k,i_2({m})}
\mleft.{\mleft[ \frac{\partial}{\partial \mb{p}}\,\mleft\{
\frac{\partial}{\partial \mb{p}'}\,f\mleft( \mb{p},\mb{p}' \mright)
\mright\}^T
\mright]}\mright|_{{\mb{p}}_{i_1({m})},{\mb{p}}_{i_2({m})},{\mb{T}}_{m},{\mb{r}}_{m}}
\mright.
\\&\phantom{{}={}}
\nonumber
{} \! + \mleft. {\mathfrak{d}}_{l,i_2({m})} {\mathfrak{d}}_{k,i_1({m})}
\mleft.{\mleft[ \frac{\partial}{\partial \mb{p}'}\,\mleft\{
\frac{\partial}{\partial \mb{p}}\,f\mleft( \mb{p},\mb{p}' \mright)
\mright\}^T
\mright]}\mright|_{{\mb{p}}_{i_1({m})},{\mb{p}}_{i_2({m})},{\mb{T}}_{m},{\mb{r}}_{m}}
\mright.
\\&\phantom{{}={}}
\nonumber
{} \! + \mleft. \vphantom{{\mathfrak{d}}_{l,i_1({m})}
{\mathfrak{d}}_{k,i_1({m})} \mleft.{\mleft[ \frac{\partial}{\partial
\mb{p}}\,\mleft\{ \frac{\partial}{\partial \mb{p}}\,f\mleft(
\mb{p},\mb{p}' \mright) \mright\}^T
\mright]}\mright|_{{\mb{p}}_{i_1({m})},{\mb{p}}_{i_2({m})},{\mb{T}}_{m},{\mb{r}}_{m}}
+ {\mathfrak{d}}_{l,i_1({m})} {\mathfrak{d}}_{k,i_2({m})}
\mleft.{\mleft[ \frac{\partial}{\partial \mb{p}}\,\mleft\{
\frac{\partial}{\partial \mb{p}'}\,f\mleft( \mb{p},\mb{p}' \mright)
\mright\}^T
\mright]}\mright|_{{\mb{p}}_{i_1({m})},{\mb{p}}_{i_2({m})},{\mb{T}}_{m},{\mb{r}}_{m}}
+ {\mathfrak{d}}_{l,i_2({m})} {\mathfrak{d}}_{k,i_1({m})}
\mleft.{\mleft[ \frac{\partial}{\partial \mb{p}'}\,\mleft\{
\frac{\partial}{\partial \mb{p}}\,f\mleft( \mb{p},\mb{p}' \mright)
\mright\}^T
\mright]}\mright|_{{\mb{p}}_{i_1({m})},{\mb{p}}_{i_2({m})},{\mb{T}}_{m},{\mb{r}}_{m}}
+ {\mathfrak{d}}_{l,i_2({m})} {\mathfrak{d}}_{k,i_2({m})}
\mleft.{\mleft[ \frac{\partial}{\partial \mb{p}'}\,\mleft\{
\frac{\partial}{\partial \mb{p}'}\,f\mleft( \mb{p},\mb{p}' \mright)
\mright\}^T
\mright]}\mright|_{{\mb{p}}_{i_1({m})},{\mb{p}}_{i_2({m})},{\mb{T}}_{m},{\mb{r}}_{m}}
} {\mathfrak{d}}_{l,i_2({m})} {\mathfrak{d}}_{k,i_2({m})}
\mleft.{\mleft[ \frac{\partial}{\partial \mb{p}'}\,\mleft\{
\frac{\partial}{\partial \mb{p}'}\,f\mleft( \mb{p},\mb{p}' \mright)
\mright\}^T
\mright]}\mright|_{{\mb{p}}_{i_1({m})},{\mb{p}}_{i_2({m})},{\mb{T}}_{m},{\mb{r}}_{m}}
\mright)
\\&\phantom{{}={}}
\nonumber
{} \! + \ldots
\\&\phantom{{}={}}
\nonumber
{} \! + \begin{pmatrix}\mb{0}_{2,2} & \mb{0}_{2,2}\\\mb{0}_{2,2} &
{\mathfrak{d}}_{k,l} {\lambda}_{k} \mb{I}_{2} \end{pmatrix}.
\end{align}
In an implementation, it is not efficient to go through all indices $k$
and $l$ as the large Hessian is sparse. Instead, we can look up all
measurements where pose $k$ is used, either in the first argument $i_1$
or in the second argument $i_2$. If pose $k$ is used in the first
argument, we only have non-zero terms if $l$ coincides with either the
first argument $i_1$ or the second argument $i_2$ (the same holds if
pose $k$ is used in the second argument). So, for each measurement where
$k$ appears in either the first or second argument, we can provide two
indices for $l$ and the corresponding second derivatives.

The remaining three terms are
\begin{align}
&\phantom{{}={}} \frac{\partial}{\partial {\lambda}_{l}}\,\mleft(
\frac{\partial}{\partial {\mb{p}}_{k}}\,L \mright)^T \nonumber \\
& = \begin{pmatrix}\mb{0}_{2}\\{\mathfrak{d}}_{l,k} {\mb{u}}_{k}
\end{pmatrix}
\end{align}
\parbox{\textwidth}{\rule{\textwidth}{0.4pt}}
\begin{align}
&\phantom{{}={}} \frac{\partial}{\partial {\mb{p}}_{l}}\,\mleft( \frac{\partial}{\partial
{\lambda}_{k}}\,L \mright)^T \nonumber \\
& = \begin{pmatrix}\mb{0}_{2}^T & {\mathfrak{d}}_{l,k} {\mb{u}}_{k}^T
\end{pmatrix}
\end{align}
\parbox{\textwidth}{\rule{\textwidth}{0.4pt}}
\begin{align}
&\phantom{{}={}} \frac{\partial}{\partial {\lambda}_{l}}\,\mleft(
\frac{\partial}{\partial {\lambda}_{k}}\,L \mright)^T \nonumber \\
& = 0.
\end{align}
With these four terms, a block $(k, l)$ of the Hessian for the total
cost function is given by
\begin{align}
& \begin{pmatrix}\frac{\partial}{\partial {\mb{p}}_{l}}\,\mleft(
\frac{\partial}{\partial {\mb{p}}_{k}}\,L \mright)^T &
\frac{\partial}{\partial {\lambda}_{l}}\,\mleft(
\frac{\partial}{\partial {\mb{p}}_{k}}\,L
\mright)^T\\\frac{\partial}{\partial {\mb{p}}_{l}}\,\mleft(
\frac{\partial}{\partial {\lambda}_{k}}\,L \mright)^T &
\frac{\partial}{\partial {\lambda}_{l}}\,\mleft(
\frac{\partial}{\partial {\lambda}_{k}}\,L \mright)^T\end{pmatrix}.
\end{align}

\subsection{Initial Values for Lagrange Multipliers}

%
The initial total state vector of our Lagrange-Newton problem includes
the initial robot poses, but also the Lagrange multipliers for which no
initial values are known. \citet[sec.~4.1.1]{nn_Izmailov14} present an
equation to determine initial estimates of the Lagrange multipliers from
the remaining initial state (under the assumption that the initial state
is close to the solution). We apply their equation for the case at hand
to determine the initial Lagrange multipliers from the initial robot
poses.

In the following, we describe vectors and matrices by their elements,
with indices $i$ and $j$ running from $2$ to $N$ (since pose $1$ is kept
fixed and therefore excluded from the update).

The inner constraint term (without the Lagrange multiplier) is
\begin{align}
{l}_{i}
&=
\frac{1}{2}\, \mleft( {\mb{u}}_{i}^T {\mb{u}}_{i} - 1 \mright).
\end{align}
Note that since the inner constraint term only depends on ${\mb{u}}_{i}$
but not ${\mb{x}}_{i}$, all parts of the equation only relate to these
terms.

The derivative of the inner constraint term is
\begin{align}
\frac{\partial}{\partial {\mb{u}}_{j}}\,{l}_{i}
&=
{\mathfrak{d}}_{i,j} {\mb{u}}_{i}^T.
\end{align}
We put all Lagrange multipliers into a vector and all derivatives of the
inner constraint term into a block matrix
\begin{align}
\bs{\lambda}
&=
{\begin{pmatrix}{\lambda}_{i}\end{pmatrix}}_{i, 1}\\
{\begin{pmatrix}\frac{\partial}{\partial
{\mb{u}}_{j}}\,{l}_{i}\end{pmatrix}}_{i, j} & = {\begin{pmatrix}{\mathfrak{d}}_{i,j} {\mb{u}}_{j}^T \end{pmatrix}}_{i,
j}
\end{align}
and apply the equation mentioned above, where $F$ is taken from
\eqref{eq_F}:
\begin{align}
&\phantom{{}={}} \bs{\lambda} \nonumber \\
& = - \mleft( {\begin{pmatrix}\frac{\partial}{\partial
{\mb{u}}_{j}}\,{l}_{i}\end{pmatrix}}_{i, j}
{\begin{pmatrix}\frac{\partial}{\partial
{\mb{u}}_{j}}\,{l}_{i}\end{pmatrix}}_{i, j}^T \mright)^{-1}
{\begin{pmatrix}\frac{\partial}{\partial
{\mb{u}}_{j}}\,{l}_{i}\end{pmatrix}}_{i, j} {\begin{pmatrix}\mleft(
\frac{\partial}{\partial {\mb{u}}_{i}}\,F \mright)^T\end{pmatrix}}_{i,
1}\\
&\label{eq_lambda_init2}
 = - \mleft( {\begin{pmatrix}{\mathfrak{d}}_{i,j} {\mb{u}}_{j}^T
\end{pmatrix}}_{i, j} {\begin{pmatrix}{\mathfrak{d}}_{i,j}
{\mb{u}}_{j}^T \end{pmatrix}}_{i, j}^T \mright)^{-1}
{\begin{pmatrix}{\mathfrak{d}}_{i,j} {\mb{u}}_{j}^T \end{pmatrix}}_{i,
j} {\begin{pmatrix}\mleft( \frac{\partial}{\partial {\mb{u}}_{i}}\,F
\mright)^T\end{pmatrix}}_{i, 1}\\
& = - {\begin{pmatrix}{\mathfrak{d}}_{i,j} {\mb{u}}_{j}^T \end{pmatrix}}_{i,
j} {\begin{pmatrix}\mleft( \frac{\partial}{\partial {\mb{u}}_{i}}\,F
\mright)^T\end{pmatrix}}_{i, 1}\\
& = - {\begin{pmatrix}{\mb{u}}_{i}^T \mleft( \frac{\partial}{\partial
{\mb{u}}_{i}}\,F \mright)^T \end{pmatrix}}_{i, 1},
\end{align}
or, for single elements:
\begin{align}
\label{eq_la_init}
{\lambda}_{i}
&=
- {\mb{u}}_{i}^T \mleft( \frac{\partial}{\partial {\mb{u}}_{i}}\,F
\mright)^T.
\end{align}
Since all ${\mb{u}}_{i}$ are initially unit-length vectors, the first
factor (and its inverse) in \eqref{eq_lambda_init2} is a unit matrix and
was therefore omitted. The derivative of $F$ is evaluated for the
initial values of the poses ${\mb{p}}_{i}$.

\subsection{Newton Method}\label{sec_newton}

%
The gradient blocks derived in section \ref{sec_total_gradient} are
combined in the total gradient vector $\mb{g}$. The Hessian blocks
derived in section \ref{sec_total_hessian} are combined in the total
Hessian $\mb{H}$. We then use Newton's method to derive the Newton step
$\bs{\Delta}\mb{s}$:
\begin{align}
\label{eq_newton_step}
\mb{H} \bs{\Delta}\mb{s}
&=
- \mb{g}.
\end{align}
Note that $g$, $\mb{H}$ and $\bs{\Delta}\mb{s}$ do not include the first
pose (see section \ref{sec_exp}).

The Newton step $\bs{\Delta}\mb{s}$ determines the direction of a
line-search which finally gives the change of the state (comprising both
poses and Lagrange multipliers). In the Lagrange-Newton method, the
criterion for the line-search cannot just be the Lagrangian (as we
approach a saddle point). Instead, we use an ``Augmented Lagrangian
Function'' which adds the Lagrangian at the current state and the
(scaled) L1-norm of the constraints (which is supposed to mitigate slow
convergence due to the ``Maratos effect''); see
\citet[sec.~5.5.1]{nn_Biegler10} and \citet[eq.~(6.87)]{nn_Izmailov14}.

If no improvement is achieved in the line-search, we apply a form of the
Levenberg-Marquardt method where a diagonal regularization matrix
$\mb{R}$ is added to the Hessian
\begin{align}
\mleft( \mb{H} + \mb{R} \mright) \bs{\Delta}\mb{s}
&=
- \mb{g}.
\end{align}
As suggested by \citet[eq.~(5.12)]{nn_Biegler10}, the matrix $\mb{R}$
comprises the following blocks on the main diagonal:
\begin{align}
& \begin{pmatrix}\eta_W & 0 & 0 & 0 & 0\\0 & \eta_W & 0 & 0 & 0\\0 & 0 &
\eta_W & 0 & 0\\0 & 0 & 0 & \eta_W & 0\\0 & 0 & 0 & 0 & -
\eta_A\end{pmatrix},
\end{align}
where $\eta_W \ge 0$ is related to the core of the bordered Hessian and
$\eta_A \ge 0$ to its border. We currently use a crude method where both
factors are increased (in a zigzag pattern through both values) until a
line-search is successful (this method may be accelerated if those
combinations are tested earlier where only one factor is zero).

If the line-search still fails, a short step is executed in the
line-search direction regardless; this is an emergency measure (and
rarely happens in the simulations).

If either $\mleft\| \bs{\Delta}\mb{s} \mright\|$ or $\mleft\| \mb{g}
\mright\|$ fall below given limits, the method terminates \citep[similar
to][algorithm~5.1]{nn_Biegler10}.

\subsection{Discontinuity Handling for Home-Vector Error}

%
The home-vector error depends on the normalized distance vector between
the two poses involved in the measurement. This leads to a
discontinuity: For small distances between the two poses, even small
changes in the distance vector may result in large changes in the
angular terms. We therefore introduced a threshold in the distance
between two poses. If the distance is smaller than the threshold, the
measurement is ignored in the Lagrange-Newton step. (The same mechanism
is also applied for the distance error, but it may not be required
there; the distance error is in any case not used in the simulation.)

\section{Experiments}\label{sec_exp}

%
We performed a preliminary, non-systematic simulation in Octave (largely
compatible with Matlab) to test our approach.\footnote{Available from
\url{https://www.ti.uni-bielefeld.de/html/downloads}. Note that the
implementation is not tuned for speed.} The simulation assumes a robot
with differential steering. The robot travels along three lanes which
are almost straight according to its odometry measurements, but in fact
are curved due to differences in the actual wheel speeds. The lanes are
split into segments. Odometry estimates (and view locations, see below)
are stored at all start/end points of the lane segments. Odometry
estimates between subsequent points on the lane are derived from an
Euler (first-order) approximation of the kinematics differential
equation (with multiple steps between the lane points). The odometry
covariance matrix (in robot coordinates of the first pose) is obtained
by error propagation; it starts from a zero covariance matrix at one
lane point, and the iteratively updated covariance matrix is stored at
the next point. There are no odometry measurements between the lanes.

It is assumed that the robot also collects panoramic images at the lane
points. Homing measurements are emulated without image processing by
adding noise to the true home-vector and compass angles. When a
panoramic image was collected at a lane point, homing measurements are
performed for three images from nearby points on the previous lane (only
two images at the start and the end of the lane). Note that only these
homing measurements can establish the spatial relations between the
lanes as there are no inter-lane odometry measurements. This corresponds
to a situation in a cleaning scenario where the robot only uses a short
segment of its current trajectory and a short segment from a previously
built map to establish a geometric relation which then e.g. can be used
to travel in a fixed distance from the previous segment.

The state (comprising poses and Lagrange multipliers) is initialized
with the {\em estimated} poses and the initial Lagrange multipliers
determined from \eqref{eq_la_init}. The state update considers
translation and rotation error as well as home-vector and the compass
error; the distance error is not used here. Hessian and gradient are
computed and the Lagrange-Newton descent determines a state vector
change (Octave's \texttt{linsolve} function is used which is also
suitable for sparse matrices); line-search and, if necessary,
Levenberg-Marquardt regularization are applied as described in section
\ref{sec_newton}.

The Lagrange-Newton step is not applied to the first pose of the first
lane in order to bind the state to the world coordinate system.
Moreover, since the solution would otherwise not be unique, excluding
the first view may be necessary for the method to converge in general
(not tested). (In the cleaning robot scenario, it would likely not be
the first pose that is fixed, but the current pose of the robot.)

\begin{figure}[tp]
\begin{center}
\includegraphics[width=0.75\textwidth]{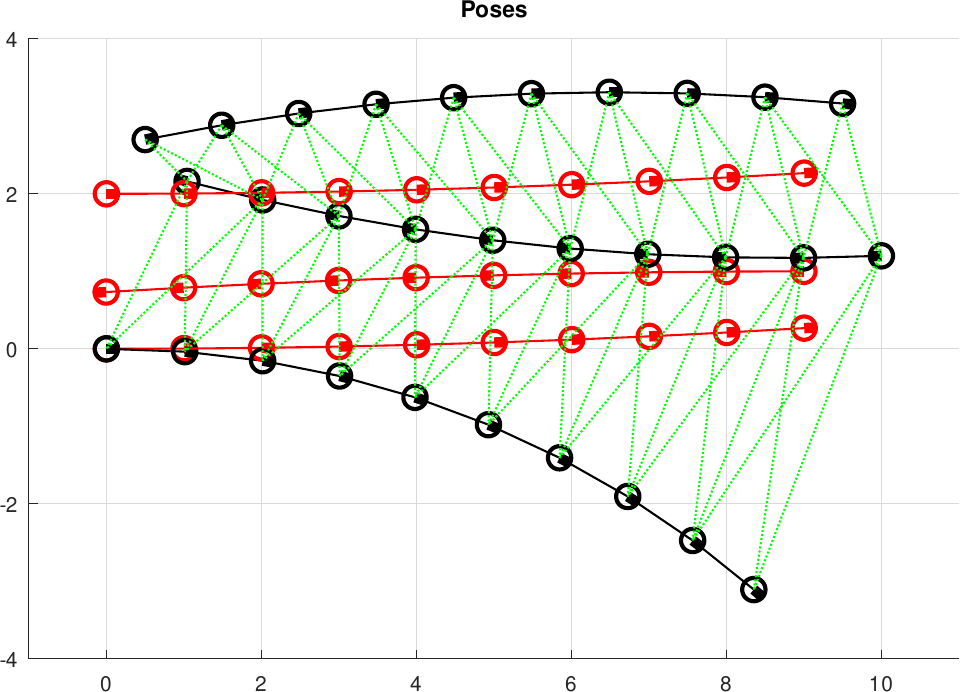}\\[5mm]
\includegraphics[width=0.75\textwidth]{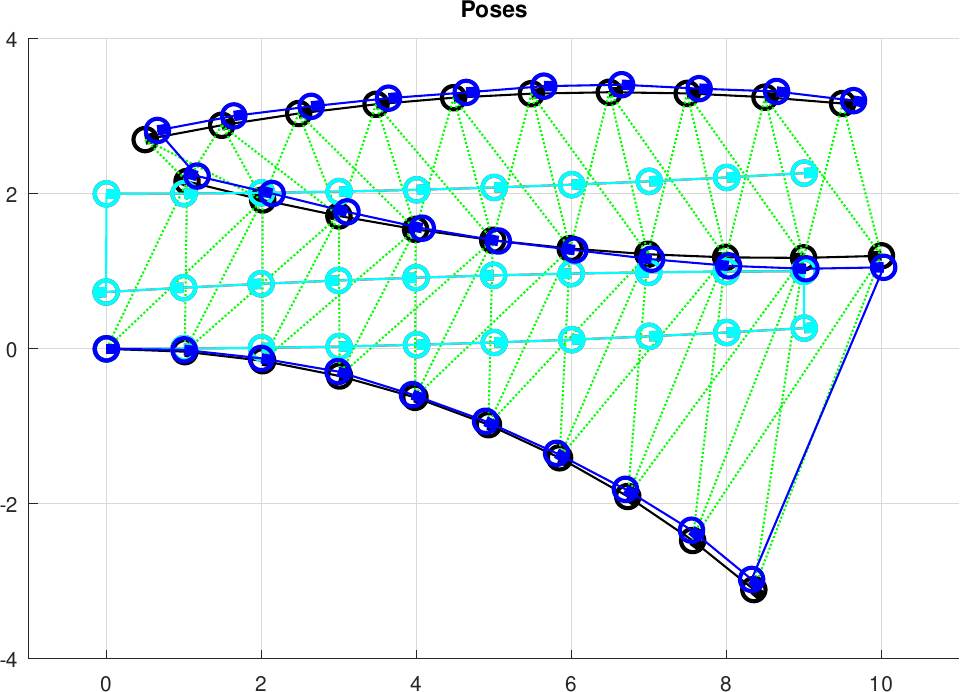}

\caption{Simulation experiment. Robot poses are shown by circles
with a bar indicating the forward direction. \textbf{Top}:
Estimated poses (red), true poses (black), and pairs of poses used
in homing measurements (dotted green lines). The first lane starts
at the origin, subsequent lanes are shifted
upwards. \textbf{Bottom}: Result of the Lagrange-Newton descent
overlaid on the top figure. Initial state (cyan), final state
(blue), pairs of poses used in homing measurements (dotted green
lines). (Inter-lane connections in the initial and final states
should be ignored.)}

\label{fig_simulation}
\end{center}
\end{figure}

In most runs, the state approaches the {\em true} poses. Regularization
of the Hessian is only rarely necessary. Note that the final state is
always a compromise between the conflicting odometry and homing
measurements, so an exact match between the solution and the true poses
cannot be expected. Typically, the total costs in the found solution are
better than the total costs in the true poses, since errors in the
homing measurements are not corrected in the true poses. Figure
\ref{fig_simulation} shows a successful run (termination criterion
reached after 10 iterations). However, in a few runs, the method
diverges and produces totally defective solutions. The reasons for this
need to be explored.

\section{Conclusions}\label{sec_conclusions}

%
Our preliminary experiments demonstrate that the suggested method of
expressing rotational components through orientation vectors --- which
are pushed towards unit length by Lagrange constraints --- is feasible.
It eliminates the need to transform variations between manifolds in each
step which is required in the Gauss-Newton approach. An obstacle to an
application are the numerical instabilities which occur in a few runs
depending on the random errors in the measurements.

We noticed that the initialization of the Lagrange multipliers using
\eqref{eq_la_init} seems to be important to achieve fast convergence.
Using initialization values far from the solution delays convergence as
first the method approaches the constraint before starting to reduce the
cost functions. In some runs we observed numerical instabilities if
unsuitable initial values were used.

We tested three different types of rotational cost functions with
increasing complexity (see section \ref{sec_cost_genrot}): the first
form \eqref{eq_first_form} with $t_1 = 1$, the first form
\eqref{eq_first_form} with $t_1 = 0$, and the second form
\eqref{eq_second_form}. The simulations show no fundamental differences
in performance; we just had the impression that the second form may
require Levenberg-Marquardt steps more often, but this needs to be
studied in detail. We cautiously conclude that it is (1) not necessary
to make the cost function independent of constraint violations and (2)
not even necessary to guarantee non-negative costs when the constraints
are violated. It would therefore be justified to just use the simplest
form of a cost function which has the desired effect when the
constraints are fulfilled. Note that the translation error just
guarantees non-negativity but not independence of constraint violations;
we didn't explore a version fulfilling the strongest requirement for
this cost function.

The methods currently used for line-search and Levenberg-Marquardt
regularization are not very sophisticated since they were simplified
from the methods suggested in the literature. Also there is no mechanism
that modifies the steps far from the solution to accelerate the
convergence.

We only tested the method for the setup shown in figure
\ref{fig_simulation}; extensive tests with different and larger graphs
are required. Moreover, none of the method's parameters were
systematically varied: We didn't explore the influence of the weight
factor for rotational cost functions $\gamma$ (simulations use $\gamma =
1$); it is also not clear whether the effect of $\gamma$ is actually
removed by the Newton method. We just superficially looked for suitable
weight factors for the L1-term in the augmented Lagrangian, for the sets
of the two weight factor in the Levenberg-Marquardt regularization, and
for the resolution and range of the line-search (see section
\ref{sec_newton}).

Finally, we raise the question whether the method can be extended to
angles in {\em three} dimensions. Presently, our derivation and
simulation are restricted to robot movements in the plane. In this case,
each angle is expressed by a two-dimensional orientation vector; all
rotations are implicitly related to the axis perpendicular to the plane.
The lengths of all orientation vectors are pushed towards unity by
Lagrange constraints. A two-by-two orientation matrix is formed from an
orientation vector to express rotations and projections. In {\em three}
dimensions, we expect that two vectors are sufficient to express each
angle. These would be complemented by a third vector to a Cartesian
right-handed coordinate system; the axes of this coordinate system would
form the orientation matrix. The Lagrange constraint would enforce that
the matrix holding the two orientation vectors in its columns would be
semi-orthogonal.

\section*{Document Changes}

%
24 Jan 2024: first release

\end{document}